\documentclass[10pt,twocolumn]{article}

\usepackage[utf8]{inputenc}
\usepackage[T1]{fontenc}
\usepackage[top=18mm,bottom=19mm,left=17mm,right=17mm,columnsep=6mm]{geometry}
\usepackage{amsmath,amssymb,amsfonts,amsthm}
\usepackage{graphicx}
\usepackage{booktabs}
\usepackage{multirow}
\usepackage{algorithm}
\usepackage{cite}
\usepackage{xcolor}
\usepackage{microtype}
\usepackage{hyperref}

\graphicspath{{../figures/}{./figures/}{./}}

\hypersetup{
    colorlinks=true,
    linkcolor=blue!70!black,
    citecolor=blue!70!black,
    urlcolor=blue!70!black,
    pdftitle={Fractal Neural Synthesis: Constant-Storage Procedural Weight Derivation via Complex Quadratic Boundary Dynamics},
    pdfauthor={Volkan Dagli, Zerrin Dagli, Daghan Dagli}
}

\title{\textbf{\Large Fractal Neural Synthesis: Constant-Storage Procedural Weight Derivation via Complex Quadratic Boundary Dynamics}}

\author{
    \textbf{Volkan Da\u{g}l\i}\textsuperscript{1,2,*}\quad
    \textbf{Zerrin Da\u{g}l\i}\textsuperscript{3,4}\quad
    \textbf{Da\u{g}han Da\u{g}l\i}\textsuperscript{5} \\[5pt]
    \footnotesize\textsuperscript{1}\textit{Anadolu University, Eski\c{s}ehir, T\"{u}rkiye} \;\textbullet\;
    \textsuperscript{2}\textit{ITouch Systems, Cukurova University Technopark, Adana, T\"{u}rkiye} \\
    \footnotesize\textsuperscript{3}\textit{Mersin University, Mersin, T\"{u}rkiye} \;\textbullet\;
    \textsuperscript{4}\textit{Yusuf Kalkavan Anatolian High School, Ministry of National Education, Mersin, T\"{u}rkiye} \\
    \footnotesize\textsuperscript{5}\textit{Toros Science High School, MEV (Toros University), Mersin, T\"{u}rkiye} \\[3pt]
    \footnotesize ORCID: V.~Da\u{g}l\i\ 0009-0000-1587-8703 \;\textbullet\;
    Z.~Da\u{g}l\i\ 0000-0001-9490-6425 \;\textbullet\;
    D.~Da\u{g}l\i\ 0009-0003-2492-8313 \\
    \footnotesize\textsuperscript{*}Corresponding author: \texttt{ask@answerr.me} \;\textbullet\;
    Code \& Data Archive: \url{https://doi.org/10.5281/zenodo.22774934}
}

\date{November 2026}

\begin{document}

\maketitle

\begin{abstract}
Conventional neural networks store synaptic parameters as explicit tensors whose persistent Flash footprint scales linearly ($O(W)$) with parameter count, constraining resource-limited edge microcontrollers. We formulate and systematically evaluate \textbf{Fractal Neural Synthesis (FNS)}, a procedural weight derivation architecture that replaces hidden-layer matrices with a constant $24\text{-byte}$ coordinate seed $\Theta = (c_x, c_y, \ell) \in \mathbb{F}_{64}^3$ anchored near the complex quadratic bifurcation boundary $\partial\mathcal{M}$ of $f_c(z)=z^2+c$. We introduce a large-bailout ($R=256$) smooth escape kernel, four zero-overhead geometric sampling apertures---Cartesian Grid, coprime $3\times 3$ Rule-of-Thirds ($\mathbb{Z}/9\mathbb{Z}$), $\phi$-Vogel spiral, and multi-scale Harmonic Tripod---and prove conditional Kolmogorov ($K_{\mathrm{Kolm}} \le 192+c_0$ bits) and Shannon entropy bounds. Across magnifications $\zeta\in[10^0,10^8]$, $C^1$ Fourier and lattice-noise bases collapse to post-standardization rank $\approx 2.25$ and SHA-256 PRNG exhibits avalanche discontinuity ($\rho\approx 0.019$, $50.6\%$ micro-step retention), whereas $\partial\mathcal{M}$ uniquely sustains both rank ($\approx 10.0$) and local continuity ($\rho\in[0.81,0.97]$, $89.2\%$ retention). Under leak-free 10-fold CV across four zero-scaler-byte non-linear benchmarks, FNS with an INT8 readout surpasses $0\text{-hidden}$ Linear-INT8 by $+3.6$ to $+10.8$ pp while delivering $10.7\times$--$129\times$ total Flash reduction ($725\times$--$2{,}773\times$ hidden reduction) relative to converged Float32 MLPs ($90.30\%$ Ionosphere in $285\text{ B}$; $97.22\%$ Optical Digits in $2{,}598\text{ B}$; $97.90\%$ PenDigits in $2{,}598\text{ B}$; $87.00\%$ Satimage in $1{,}570\text{ B}$).
\end{abstract}

\vspace{0.15cm}


\section{Introduction}
\label{sec:intro}

Modern neural networks achieve strong empirical generalization by storing every synaptic weight as an explicit numerical scalar in dense tensor arrays \cite{vaswani2017attention}, incurring linear $O(W)$ persistent memory growth in the total scalar parameter count $W$. While manageable on datacenter accelerators, this linear footprint creates a severe non-volatile Flash bottleneck on resource-constrained edge microcontrollers, embedded telemetry nodes, and autonomous sensors where on-chip Flash ROM is tightly bounded ($32\text{--}256\text{ KB}$) and static memory reads dominate system energy \cite{lin2020mcunet,stmicro2021stm32l}. Compression techniques---including post-training quantization, 1.58-bit ternary models \cite{ma2024bitnet}, sparse subnetwork pruning \cite{frankle2019lottery}, structured kernel transforms \cite{le2013fastfood}, and hash-bucket weight sharing \cite{chen2015hashednets}---reduce the constant bit prefactor per weight, yet retain $O(W)$ or $O(B_{\mathrm{hash}})$ persistent storage scaling.

Biological nervous systems adopt a contrasting developmental strategy: compact genomic encodings orchestrate far more synaptic junctions than explicit point-by-point storage could accommodate by relying on recursive, self-similar morphogenetic rules. In one-dimensional complex dynamics, the connectedness locus $\mathcal{M} \subset \mathbb{C}$ of the monic quadratic family $f_c(z) = z^2 + c$ \cite{mandelbrot1980fractal,mandelbrot1982fractal,douady1984etude,peitgen1986beauty} offers an analytic extreme of procedural structure: a lookup-free recurrence generates a non-rectifiable bifurcation boundary $\partial\mathcal{M}$ whose local Hausdorff dimension is $\dim_H(\partial\mathcal{M} \cap U) = 2$ for every open set $U$ intersecting $\partial\mathcal{M}$ \cite{shishikura1998hausdorff}.

In this paper, we formulate and systematically evaluate \textbf{Fractal Neural Synthesis (FNS)}, a procedural weight derivation framework that replaces explicit hidden-layer storage with a constant $24\text{-byte}$ IEEE-754 Float64 coordinate seed $\Theta = (c_x, c_y, \ell) \in \mathbb{F}_{64}^3$ anchored near $\partial\mathcal{M}$. Consistent with fundamental Shannon and Kolmogorov information limits \cite{shannon1948mathematical,li2019kolmogorov}, we do not claim to compress arbitrary high-entropy frontier language models into $24\text{ B}$. Rather, we address three concrete representation-learning and TinyML systems questions: \textit{(i)~Under leak-free evaluation on genuinely non-linear, zero-scaler-byte real-world benchmarks, how does FNS compare against quantized linear classifiers, seeded-PRNG random features, hashed reservoirs, and converged quantized MLPs? (ii)~What measurable structural properties separate $\partial\mathcal{M}$ from cryptographic PRNGs, smooth Fourier sums, and fixed-lattice value noise? (iii)~How do geometric sampling apertures and gradient-free optimizers govern spectral conditioning and zero-readout synthesis?}

Building substantially upon our preliminary archive \cite{dagli2026fns_zenodo} (which evaluated only 2D synthetic tasks on a uniform grid), this work makes five contributions:
\begin{enumerate}
    \item \textbf{Smooth Boundary Kernel \& Multi-Scale Generator Ablation:} We formulate a large-bailout ($R=256$) smooth escape-time kernel and evaluate $\partial\mathcal{M}$ against SHA-256 PRNG, smooth 2D Fourier bases, and fixed-lattice value noise across eight magnification decades ($\zeta = 10^\ell$, $\ell \in \{0,\dots,8\}$). We demonstrate empirically that smooth $C^1$ generators collapse after standardization to an effective matrix rank of $\approx 2.25$ (verifying a planar Taylor-ramp rank bound), whereas SHA-256 PRNG exhibits complete cryptographic avalanche ($\rho \approx 0.019$, collapsing to $50.60\%$ accuracy under a $1\%$ micro-step). By contrast, $\partial\mathcal{M}$ uniquely combines sustained post-standardization rank ($\approx 10.0$), local continuity ($\rho \in [0.81, 0.97]$), and $89.15\%$ micro-step stability with zero lookup tables.
    \item \textbf{Four Geometric Sampling Apertures:} We formalize and compare four zero-overhead coordinate stencils---a baseline Cartesian Grid, a $3\times 3$ Rule-of-Thirds nodal stencil aligned with $\mathbb{Z}/9\mathbb{Z}$ modular residue indexing and dynamically coprime sub-strides, an area-preserving $\phi$-Vogel phyllotaxis spiral \cite{vogel1979better}, and a multi-scale Harmonic Tripod ($\{0.60, 1.00, 1.60\}$ zoom cascade)---showing statistically significant accuracy improvements ($p \le 0.023$ on UCI PenDigits) and up to $+25.5\%$ relative spectral rank utilization gains at zero additional persistent storage.
    \item \textbf{Zero-Scaler-Byte Non-Linear Benchmarks \& Fair Baselines:} Using four genuinely non-linear real-world benchmarks requiring $0\text{ B}$ of feature-scaler persistence (UCI Ionosphere, Optical Digits, PenDigits, and Statlog Satimage) under unified 10-fold stratified cross-validation, we benchmark FNS transparently against $0\text{-hidden}$ Linear-INT8, $8\text{-byte}$ Seeded-PRNG RFF/ELM, fixed 2D SplitMix64 Hashed-ELM, and converged FP32/INT8/INT4 MLPs with exact layer-wise scale accounting.
    \item \textbf{Information-Theoretic Capacity Bounds \& Equal-Budget ES Scaling:} We prove conditional Kolmogorov complexity ($K_{\mathrm{Kolm}}(\mathbf{W},\mathbf{b}\mid d,H) \le 192 + c_0$ bits) and Shannon entropy bounds ($H(\mathbf{W},\mathbf{b}) \le 192$ bits) delineating FNS's System-1 operating regime, and evaluate four equal-budget ($32$ evals/gen) zero-readout multi-patch optimizers, including a Hybrid Elitist + Rank-1 ES inspired by EGGROLL \cite{sarkar2025eggroll}.
    \item \textbf{Warmed-Up Boot Synthesis \& Apples-to-Apples SRAM Profiling:} We quantify monotonic boot-time synthesis latency ($8.05\text{--}23.29\text{ ms}$) against identical pure-NumPy single-sample ($13.6\text{--}21.0\,\mu\text{s}$) and batch ($5.6\text{--}8.4\,\mu\text{s}$) SRAM inference, demonstrating parity with dense MLP forward-pass FLOPs once materialized in volatile memory.
\end{enumerate}

\section{Related Work and Comparative Positioning}
\label{sec:related}

Procedural weight derivation intersects five established research lines, summarized in Table~\ref{tab:related}.

\textbf{Indirect Encodings and Hypernetworks.} Compositional Pattern Producing Networks (CPPNs) and HyperNEAT \cite{stanley2009hyperneat} generate spatial connectivity matrices by querying an evolved coordinate neural network, while HyperNetworks \cite{ha2017hypernetworks} train an auxiliary network to predict target weights via backpropagation. Both approaches persist learned neural generators consuming kilobytes to megabytes. FractalNet \cite{larsson2017fractalnet} employs self-similar macro-routing without residual connections, but trains and stores all tensor weights explicitly. FNS eliminates learned generator parameters entirely via a lookup-free complex quadratic recurrence.

\textbf{Random Features, ELMs, Fastfood, and Hashed Reservoirs.} Extreme Learning Machines (ELMs) \cite{huang2006extreme,huang2006universal} and Random Fourier Features (RFF) \cite{rahimi2007random,sutherland2015error} pair fixed nonlinear hidden projections with closed-form $L_2$-regularized linear readouts. Fastfood \cite{le2013fastfood} compresses explicit Gaussian projections from $O(dH)$ to $O(H)$ storage via structured Walsh--Hadamard transforms, while HashedNets \cite{chen2015hashednets} hash weights into $B_{\mathrm{hash}}$ shared buckets with sign hashing during backpropagation. Importantly, fixed ELM/RFF hidden layers can themselves be regenerated at boot from an $8\text{-byte}$ PRNG seed ($O(1)$ Flash). We therefore explicitly include $8\text{-byte}$ Seeded-PRNG RFF and ELM alongside explicit RFF and untrained 2D Hashed-ELM in Section~\ref{sec:exp_real}, isolating the exact trade-offs between PRNG reservoirs and continuous $\partial\mathcal{M}$ boundary synthesis.

\textbf{Implicit Representations and TinyML Compression.} Periodic implicit neural representations such as SIREN \cite{sitzmann2020siren} compress continuous signals \textit{into} network weights, operating in the dual direction of FNS. In TinyML, MCUNet \cite{lin2020mcunet}, Lottery Ticket sparse subnetworks \cite{frankle2019lottery}, and 1.58-bit ternary weights \cite{ma2024bitnet} push explicit weight footprints toward sub-byte precision while retaining $O(W)$ asymptotic Flash scaling.

\textbf{Intrinsic Dimensionality and Backprop-Free Evolution.} Li et al.\ \cite{li2018intrinsic} showed that many learning objectives possess compact intrinsic dimensions. Black-box Evolution Strategies (OpenES) \cite{salimans2017evolution} and recent hyperscale rank-$r$ perturbation schemes such as EGGROLL \cite{sarkar2025eggroll} remove backpropagation entirely by averaging parallel low-rank perturbations, enabling pure-integer and non-differentiable training. FNS complements this paradigm by mapping search directly into low-dimensional coordinate spaces $\mathbb{F}_{64}^{P\times 3}$.

\begin{table*}[t]
\centering
\caption{Comparative positioning of Fractal Neural Synthesis (FNS) against indirect encoding, random-feature, TinyML compression, and gradient-free paradigms.}
\label{tab:related}
\resizebox{\textwidth}{!}{
\begin{tabular}{@{}llllccc@{}}
\toprule
\textbf{Paradigm / Reference} & \textbf{Persistent Storage} & \textbf{Generator / Mechanism} & \textbf{Scale-Free} & \textbf{Backprop-Free} & \textbf{Local Continuity} \\
\midrule
CPPN / HyperNEAT \cite{stanley2009hyperneat} & $O(|G_{\mathrm{CPPN}}|)$ (KB) & Evolved coordinate neural net & Bounded & Yes & Yes \\
HyperNetworks \cite{ha2017hypernetworks} & $O(|G_{\mathrm{hyper}}|)$ (MB) & Auxiliary neural generator & No & No & Yes \\
FractalNet \cite{larsson2017fractalnet} & $O(W)$ (MB) & Self-similar macro-topology only & Topology only & No & N/A \\
HashedNets / Fastfood \cite{chen2015hashednets,le2013fastfood} & $O(B_{\mathrm{hash}})$ / $O(H)$ & Hash buckets / Hadamard blocks & No & No / Mode II & No \\
Seeded-PRNG ELM / RFF \cite{huang2006extreme,rahimi2007random} & $O(1)$ ($8\text{ B}$) + $O(HC)$ head & Counter PRNG Gaussian / cosine map & Constant var & Yes (Mode II) & No ($\rho\approx 0$) \\
SIREN / INR \cite{sitzmann2020siren} & $O(W_{\mathrm{INR}})$ (KB--MB) & Sine-MLP encoding signals & Band-limited & No & Yes \\
MCUNet / BitNet \cite{lin2020mcunet,ma2024bitnet} & $O(W\cdot b/8)$ & Quantized / ternary tensors & N/A & No & N/A \\
OpenES / EGGROLL \cite{salimans2017evolution,sarkar2025eggroll} & $O(W)$ dense weights & Full- or rank-$r$ perturbations & Param space $\mathbb{R}^W$ & Yes & Yes \\
\midrule
\textbf{FNS (This Work)} & \textbf{$O(1)$ hidden ($24\text{ B}$)} + optional head & \textbf{Quadratic $\partial\mathcal{M}$ + 4 apertures} & \textbf{Yes ($\zeta \le 10^8$)} & \textbf{Yes} & \textbf{Yes ($\rho\approx 0.84$)} \\
\bottomrule
\end{tabular}
}
\end{table*}

\section{Mathematical Formulation and Theoretical Bounds}
\label{sec:math}

\begin{figure*}[t]
\centering
\includegraphics[width=\textwidth]{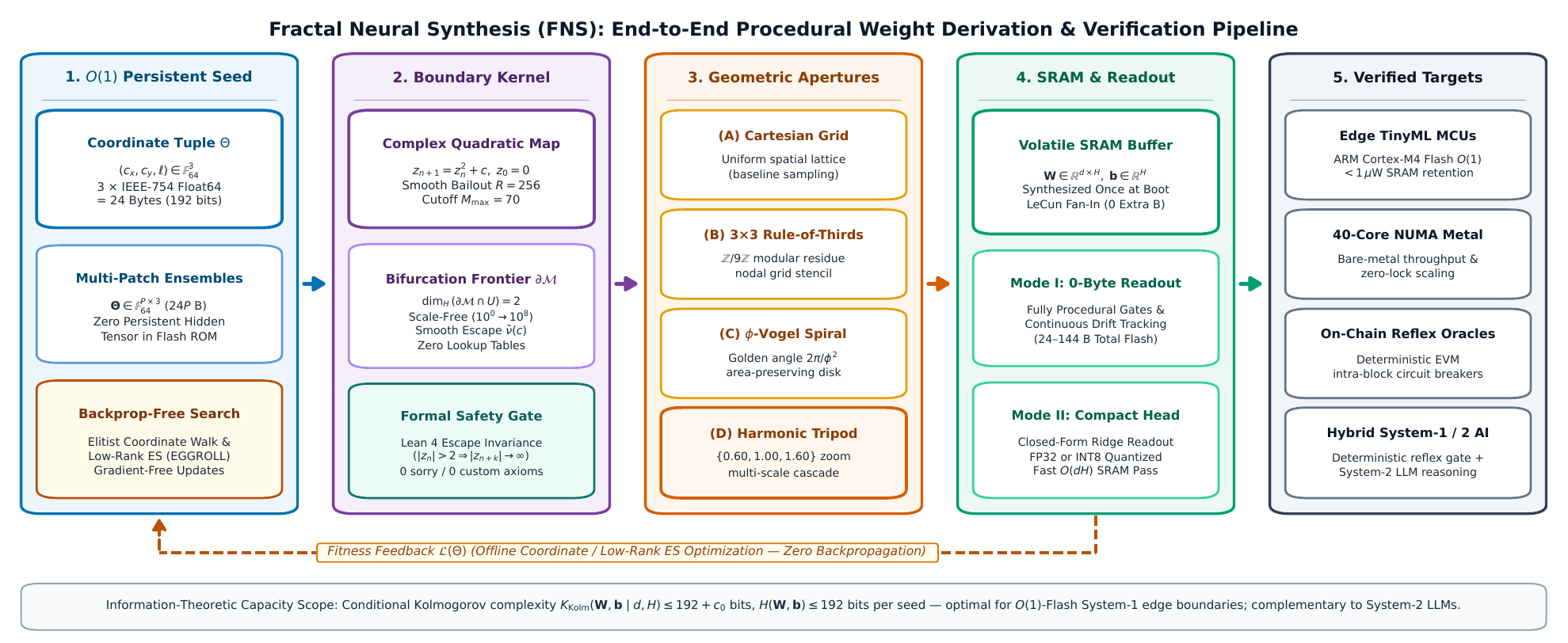}
\caption{End-to-end Fractal Neural Synthesis (FNS) pipeline. A $24\text{-byte}$ coordinate tuple $\Theta=(c_x, c_y, \ell)\in\mathbb{F}_{64}^3$ drives the lookup-free quadratic kernel $f_c(z)=z^2+c$ ($R=256$) along $\partial\mathcal{M}$, sampled through four geometric apertures into a boot-time volatile SRAM buffer supporting either 0-byte procedural Mode~I reflexes or compact FP32/INT8 Mode~II readouts.}
\label{fig:arch}
\end{figure*}

Figure~\ref{fig:arch} illustrates the end-to-end FNS pipeline from persistent $24\text{-byte}$ Flash storage through large-bailout boundary evaluation, four geometric apertures, boot-time volatile SRAM materialization, and target execution environments.

\subsection{Large-Bailout Smooth Escape-Time Kernel}
\label{sec:kernel}

Consider the monic complex quadratic recurrence initialized at the critical point $z_0 = 0$:
\begin{equation}
\label{eq:quad_map}
z_{n+1} = f_c(z_n) = z_n^2 + c, \qquad c = c_x + i c_y \in \mathbb{C}.
\end{equation}
The complex parameter plane partitions disjointly as $\mathbb{C} = \mathrm{int}(\mathcal{M}) \sqcup \partial\mathcal{M} \sqcup (\mathbb{C}\setminus\mathcal{M})$, where $\mathcal{M} = \{c\in\mathbb{C} : \sup_{n\ge 0}|z_n(c)| \le 2\}$ comprises interior hyperbolic components and the bifurcation boundary $\partial\mathcal{M}$ \cite{douady1984etude,milnor2006dynamics}. For bounded evaluation up to iteration cutoff $M_{\max}$ (default $M_{\max}=70$) with large bailout radius $R = 256.0$, we define the escape-time index without sentinel collision:
\begin{equation}
\label{eq:escape_idx}
N_{\mathrm{esc}}(c) =
\begin{cases}
\min\bigl\{n \le M_{\max} : |z_n(c)| > R\bigr\}, & \text{if escaped}, \\
\infty, & \text{otherwise},
\end{cases}
\end{equation}
with $n \in \{1,\dots,M_{\max}\}$. Two deterministic fields follow directly from Eq.~\eqref{eq:escape_idx}:
\begin{itemize}
    \item \textbf{Binary Bounded Indicator} $I(c) = \mathbb{I}[N_{\mathrm{esc}}(c) = \infty] \in \{0,1\}$, used for spatial area integration in Mode~I quadrant boolean units;
    \item \textbf{Normalized Smooth Escape-Time Potential} $\tilde{\nu}(c) \in [0,1]$, set to $\tilde{\nu}(c)=1.0$ if $N_{\mathrm{esc}}(c)=\infty$, and otherwise defined via the standard smooth iteration count \cite{douady1984etude,milnor2006dynamics}:
    \begin{align}
    \label{eq:smooth_nu}
    \nu(c) &= N_{\mathrm{esc}}(c) + 1 - \log_2 \log_2 |z_{N_{\mathrm{esc}}(c)}(c)|, \nonumber \\
    \tilde{\nu}(c) &= \mathrm{clip}\!\left(\frac{\nu(c)}{M_{\max}}, \; 0, \; 1 - 10^{-6}\right).
    \end{align}
\end{itemize}
Crucially, adopting a large bailout $R=256.0$ ensures $|z_n^2+c| \approx |z_n|^2$ upon escape so that $\log_2\log_2|z_{n+1}| - \log_2\log_2|z_n| = 1 + O(R^{-2})$, eliminating the macroscopic $O(1)$ jump discontinuities across escape lemniscates $\{c : |z_k(c)|=R\}$ that arise at the minimal bailout $R=2.0$. Note that $\tilde{\nu}(c)$ is a clipped, normalized monotone surrogate for $-\log_2 G(c)$, where $G(c)=\lim_{n\to\infty} 2^{-n}\ln|z_n(c)|$ is the continuous exterior Douady--Hubbard Green potential \cite{douady1984etude}.

\subsection{Bifurcation Frontier $\partial\mathcal{M}$: Parabolic vs.\ Misiurewicz Scaling}
\label{sec:horizon}

Biological motor adaptation frequently operates via directional error boundaries rather than isotropic gradient steps \cite{shadmehr2010error}. In complex parameter space, $\partial\mathcal{M}$ is nowhere dense with topological dimension $1$, yet Shishikura \cite{shishikura1998hausdorff} proved that every nonempty open set $U$ intersecting $\partial\mathcal{M}$ has maximal metric Hausdorff dimension $\dim_H(\partial\mathcal{M}\cap U) = 2$.

To prevent symbol collision with complex iterates $z_n$, let $\ell \in \mathbb{R}$ denote log-magnification, $\zeta = 10^\ell$ magnification, and $s = 10^{-\ell}$ window half-width. Escape dynamics near $\partial\mathcal{M}$ depend critically on the anchor type:
\begin{itemize}
    \item \textbf{Parabolic Pinches and Valley Anchors:} At the period-doubling parabolic neck $c = -3/4$ (multiplier $\lambda=-1$), vertical escape scales as $N_{\mathrm{esc}}(-3/4 + i\epsilon) \sim \pi/\epsilon$ \cite{peitgen1992chaos}, so a fixed cutoff $M_{\max}=70$ saturates ($I(c)\equiv 1$) already at $\zeta \approx 22$. Displacing into the Seahorse Valley exterior boundary $\Theta_0 = (-0.743643887, 0.131825904, 2.15)$ provides rich non-saturating boundary contrast at $M_{\max}=70$ for operational zooms $\ell \in [0.5, 3.5]$.
    \item \textbf{Misiurewicz Anchors ($O(\log\zeta)$ Depth):} At the pre-periodic Misiurewicz point $c_{\mathrm{Mis}} = -0.10109636384562 + 0.95628651080914i$ ($M_{3,1}$), the critical orbit lands on a repelling fixed point of multiplier $|\lambda| \approx 1.3283$ and $\partial\mathcal{M}$ is asymptotically self-similar by Tan Lei's theorem \cite{tanlei1990similarity}. For typical exterior points at distance $\sim \zeta^{-1}$ from $c_{\mathrm{Mis}}$, escape times grow logarithmically as $\Theta(\log\zeta)$, which a modest adaptive cutoff $M_{\max}(\zeta) = 70 + 40\ell$ tracks cleanly across eight decades ($\zeta \in [10^0, 10^8]$) with zero added persistent bytes.
\end{itemize}

\subsection{Four Geometric Sampling Apertures}
\label{sec:apertures}

Given a $24\text{-byte}$ coordinate seed $\Theta = (c_x, c_y, \ell) \in \mathbb{F}_{64}^3$ ($s = 10^{-\ell}$), synthesizing hidden weights $\mathbf{W}\in\mathbb{R}^{d\times H}$ and biases $\mathbf{b}\in\mathbb{R}^H$ requires $M = (d+1)H$ deterministic complex coordinates $\{C_k\}_{k=0}^{M-1}$. We define and evaluate four zero-overhead geometric apertures:

\begin{enumerate}
    \item \textbf{Cartesian Grid (Baseline):} Evaluates a uniform $\lceil\sqrt{M}\rceil\times\lceil\sqrt{M}\rceil$ lattice over $[c_x-s, c_x+s]\times[c_y-s, c_y+s]$ truncated to $M$ points. For minimal 2-input boolean gates (Experiment~1), a $128\times 128$ grid partitioned into four quadrants $(Q_1,Q_2,Q_3,Q_4)$ maps interior dark ratios $\mathbf{R}_Q \in [0,1]^4$ directly to $\boldsymbol{\theta} = (w_1,w_2,w_3,b) = 6.0(\mathbf{R}_Q - \tfrac{1}{2}\mathbf{1}_4)$ (retaining $w_1,w_2,b$).
    \item \textbf{$3\times 3$ Rule-of-Thirds Nodal Grid ($\mathbb{Z}/9\mathbb{Z}$):} Partitions the window into nine sub-basins indexed by $r_k = k \bmod 9 \in \mathbb{Z}/9\mathbb{Z}$, with macro-offsets $u_k = \lfloor r_k/3\rfloor - 1$, $v_k = (r_k \bmod 3) - 1$, and layer index $q_k = \lfloor k/9\rfloor$. Inside each $1/3\times 1/3$ basin of sub-grid width $L_{\mathrm{sub}} = \max(2, \lceil\sqrt{\lceil M/9\rceil}\rceil)$, we choose the smallest integers $p_r \ge 5, p_c \ge 7$ coprime to $L_{\mathrm{sub}}$ ($\gcd(p_r,L_{\mathrm{sub}})=\gcd(p_c,L_{\mathrm{sub}})=1$) so no sub-row or sub-column collapses:
    \begin{align}
    \label{eq:rot}
    C_k &= \left[c_x + \tfrac{2}{3}s\!\left(v_k + \tfrac{(\lfloor q_k/L_{\mathrm{sub}}\rfloor p_c + 3r_k) \bmod L_{\mathrm{sub}}}{L_{\mathrm{sub}}-1} - \tfrac{1}{2}\right)\right] \nonumber\\
        &\quad + i\left[c_y + \tfrac{2}{3}s\!\left(u_k + \tfrac{q_k p_r \bmod L_{\mathrm{sub}}}{L_{\mathrm{sub}}-1} - \tfrac{1}{2}\right)\right].
    \end{align}
    \item \textbf{Area-Preserving $\phi$-Vogel Phyllotaxis Spiral:} Distributes $M$ points at uniform areal density over the disk $D(c, s)$ using Vogel's Fermat spiral \cite{vogel1979better} and the golden angle $\alpha_\phi = 2\pi/\phi^2 \approx 2.39996\text{ rad}$ ($\phi=(1+\sqrt{5})/2$):
    \begin{align}
    \label{eq:spiral}
    C_k &= \left(c_x + s\sqrt{\frac{k+0.5}{M}}\cos(k\alpha_\phi)\right) \nonumber\\
        &\quad + i\left(c_y + s\sqrt{\frac{k+0.5}{M}}\sin(k\alpha_\phi)\right).
    \end{align}
    By Hurwitz's theorem \cite{hurwitz1891ueber}, $\phi$ is worst-approximable by rationals, preventing low-period angular clustering of consecutive indices $k, k+1$ (though parastichy alignment persists at Fibonacci column lags $F_m \in \{8,13,21,34,\dots\}$).
    \item \textbf{Multi-Scale Harmonic Tripod ($\{0.60, 1.00, 1.60\}$):} Breaks Fibonacci-lag column alignment by interleaving three concentric Vogel tiers ($t = k\bmod 3$, $j_k = \lfloor k/3\rfloor$, $M_{\mathrm{tier}} = \lceil M/3\rceil$) with zoom multipliers $\tau_t \in \{0.60, 1.00, 1.60\}$ (producing radii $s/\tau_t$) and $120^\circ$ phase shifts:
    \begin{align}
    \label{eq:tripod}
    C_k &= \left(c_x + \frac{s}{\tau_t}\sqrt{\frac{j_k+0.5}{M_{\mathrm{tier}}}}\cos\!\left(j_k\alpha_\phi + \frac{2\pi t}{3}\right)\right) \nonumber\\
        &\quad + i\left(c_y + \frac{s}{\tau_t}\sqrt{\frac{j_k+0.5}{M_{\mathrm{tier}}}}\sin\!\left(j_k\alpha_\phi + \frac{2\pi t}{3}\right)\right).
    \end{align}
\end{enumerate}

Let $\mathbf{u} = [\tilde{\nu}(C_0),\dots,\tilde{\nu}(C_{M-1})]^T \in [0,1]^M$ with population mean $\mu(\mathbf{u})$ and standard deviation $\sigma(\mathbf{u})$. Adding \textbf{0 persistent bytes}, boot-time synthesis applies per-tensor standardization ($\varepsilon = 10^{-8}$), row-major (C-order) reshaping, and LeCun fan-in variance scaling \cite{lecun1998efficient}:
\begin{align}
\label{eq:lecun}
\tilde{\mathbf{u}} &= \frac{\mathbf{u} - \mu(\mathbf{u})}{\sigma(\mathbf{u}) + \varepsilon}, \\
\mathbf{W} &= \frac{\gamma}{\sqrt{d}}\,\mathrm{reshape}_{d\times H}^{\mathrm{C}}(\tilde{\mathbf{u}}_{0:dH}), \qquad
\mathbf{b} = \frac{1}{2}\,\tilde{\mathbf{u}}_{dH:M}. \nonumber
\end{align}

\subsection{Storage Complexity, Ephemeral SRAM, and Information-Theoretic Bounds}
\label{sec:theory}

Distinguishing persistent non-volatile Flash ROM from volatile runtime SRAM clarifies FNS's systems profile:
\begin{itemize}
    \item \textbf{Persistent Flash ROM ($O(1)$ Hidden Storage):} A single seed $\Theta \in \mathbb{F}_{64}^3$ occupies exactly $24\text{ B}$ regardless of $d$ or $H$. Mode~I multi-patch networks store $24P\text{ B}$ total ($0\text{ B}$ readout). Mode~II networks store $24 + B_{\mathrm{head}}$ bytes, where $B_{\mathrm{head}} = 4(H+1)C_{\mathrm{out}}$ (FP32) or $(H+1)C_{\mathrm{out}} + 4$ (symmetric INT8 + one 4-byte Float32 scale), with $C_{\mathrm{out}}=1$ for binary and $C_{\mathrm{out}}=C$ for multi-class tasks.
    \item \textbf{Ephemeral Volatile SRAM Scratchpad:} Synthesis executes strictly once at device power-on, populating $(\mathbf{W},\mathbf{b})$ in volatile SRAM. Runtime inferences then execute standard dot products with zero escape-time iterations; during MCU duty-cycle sleep, modern ARM Cortex-M4 standby retention holds kilobytes of SRAM below $1\,\mu\text{W}$ \cite{stmicro2021stm32l}.
\end{itemize}

\newtheorem{proposition}{Proposition}
\newtheorem{remark}{Remark}

\begin{proposition}[Information-Theoretic Capacity Ceiling]
\label{prop:capacity}
Let $\mathcal{F}_{\mathrm{FNS}} : \mathbb{F}_{64}^3 \to \mathbb{F}_{64}^{d\times H}\times\mathbb{F}_{64}^H$ denote the deterministic FNS synthesis map of Eqs.~\eqref{eq:quad_map}--\eqref{eq:lecun} for fixed dimensions $(d,H)$, cutoff $M_{\max}$, aperture $\mathcal{A}$, and fixed program length $c_0$ bits. Then:
\begin{enumerate}
    \item \textbf{Conditional Kolmogorov Complexity:} Every synthesized hidden tensor satisfies $K_{\mathrm{Kolm}}(\mathbf{W},\mathbf{b}\mid d, H, M_{\max}, \mathcal{A}) \le 192 + c_0$ bits \cite{li2019kolmogorov}, independent of $d$ and $H$.
    \item \textbf{Image Cardinality \& Shannon Entropy:} The image contains strictly fewer than $2^{192}$ distinct hidden tensors, $|\mathrm{Im}(\mathcal{F}_{\mathrm{FNS}})| < 2^{192}$. For any distribution over finite seeds $\Theta \in \mathbb{F}_{64}^3$, the Shannon entropy satisfies $H(\mathbf{W},\mathbf{b}) \le H(\Theta) \le 192$ bits \cite{shannon1948mathematical}.
\end{enumerate}
\end{proposition}

\begin{proof}
(1) Given $(d, H, M_{\max}, \mathcal{A})$ as side information, a prefix-free program of constant length $c_0$ parses the $192\text{-bit}$ IEEE-754 representation of $\Theta$ and deterministically executes Eqs.~\eqref{eq:quad_map}--\eqref{eq:lecun} to emit $(\mathbf{W},\mathbf{b})$, establishing $K_{\mathrm{Kolm}}(\mathbf{W},\mathbf{b}\mid d,H,M_{\max},\mathcal{A}) \le 192 + c_0$. (2) Because $\mathbb{F}_{64}^3$ has at most $2^{192}$ bit patterns (strictly fewer excluding \texttt{NaN}/\texttt{Inf}) and $\mathcal{F}_{\mathrm{FNS}}$ is deterministic, $|\mathrm{Im}(\mathcal{F}_{\mathrm{FNS}})| \le |\mathbb{F}_{64}^3| < 2^{192}$; by the data processing inequality for discrete entropy, $H(\mathcal{F}_{\mathrm{FNS}}(\Theta)) \le H(\Theta) \le \log_2|\mathbb{F}_{64}^3| \le 192$ bits.
\end{proof}

\begin{remark}[Mode II Expressivity and Open Density Question]
\label{rem:expressivity}
In Mode~II, adding a trainable linear head $\mathbf{W}_{\mathrm{out}} \in \mathbb{R}^{(H+1)\times C_{\mathrm{out}}}$ lifts the $192\text{-bit}$ hidden ceiling to $O(HC_{\mathrm{out}})$ total storage while keeping hidden Flash $O(1)$. By classical ridge-function theory \cite{cybenko1989approximation,leshno1993multilayer,huang2006universal}, universal approximation on compact subsets $K \subset \mathbb{R}^d$ holds if the hidden feature dictionary $\{1, \tanh(\mathbf{w}_j^T\mathbf{x}+b_j)\}_{j=1}^H$ spans a dense subspace of $C(K)$ as $H\to\infty$. Because FNS rows are deterministic evaluations of $\tilde{\nu}$ rather than i.i.d.\ draws from a full-support measure, FNS does not automatically inherit probabilistic ELM/RFF universality theorems, and full matrix rank $\mathrm{rank}_{\mathrm{eff}}(\mathbf{W})$ alone does not imply $C(K)$-density. Proving analytical density of deterministic $\partial\mathcal{M}$-generated dictionaries remains an open theoretical question; here we assess conditioning empirically via effective spectral rank \cite{roy2007effective} and held-out cross-validation.
\end{remark}

Proposition~\ref{prop:capacity} and Remark~\ref{rem:expressivity} define FNS's epistemic scope: consistent with intrinsic-dimension theory \cite{li2018intrinsic} and No Free Lunch bounds \cite{wolpert1997no}, FNS targets compact, deterministic System-1 classifiers and edge controllers rather than high-entropy System-2 language models.

\section{Backprop-Free Synthesis and Optimization Algorithms}
\label{sec:alg}

FNS trains without computing backward gradients through hidden weight tensors under two operational regimes unified in Algorithm~\ref{alg:fns}:

\textbf{Mode I (Zero-Readout Coordinate Search).} For boolean gates ($24\text{ B}$/unit), 2-layer XOR ($72\text{ B}$), and compact zero-readout classifiers ($48\text{--}144\text{ B}$), all weights are synthesized from coordinate matrices $\boldsymbol{\Theta}\in\mathbb{F}_{64}^{P\times 3}$. We compare four gradient-free optimizers under a strict equal evaluation budget ($N_{\mathrm{pop}}$ candidates/generation):
\begin{itemize}
    \item \textit{Elitist Coordinate Walk:} Retains the best anchor verbatim and mutates $(c_x, c_y)$ with scale-coupled variance $\sigma_c^2 s^2$ and $\ell$ with $\sigma_\ell^2$.
    \item \textit{Antithetic Full-Rank ES \cite{salimans2017evolution}:} Samples $N_{\mathrm{pop}}/2$ i.i.d.\ matrices $[\boldsymbol{\epsilon}_i]_{p,j} \sim \mathcal{N}(0,1)$, evaluates antithetic pairs $\boldsymbol{\Theta} \pm \boldsymbol{\sigma}\odot\boldsymbol{\epsilon}_i$, and updates via rank-shaped fitness differences.
    \item \textit{EGGROLL-Inspired Rank-1 ES \cite{sarkar2025eggroll}:} Replaces full-rank perturbations with rank-1 outer products $\mathbf{E}_i = \mathbf{u}_i\mathbf{v}_i^T$ ($\mathbf{u}_i \sim \mathcal{N}(\mathbf{0}, \mathbf{I}_P)$, $\mathbf{v}_i \sim \mathcal{N}(\mathbf{0}, \mathbf{I}_3)$), matching entrywise unit variance while cutting per-worker noise dimension from $3P$ to $P+3$.
    \item \textit{Hybrid Elitist + Rank-1 ES:} Anchors the rank-1 ES iterate $\boldsymbol{\Theta}_{\mathrm{ES}}$ toward the best discovered boundary state $\boldsymbol{\Theta}^*$ via convex pull $\boldsymbol{\Theta} \leftarrow 0.55\,\boldsymbol{\Theta}^* + 0.45\,\boldsymbol{\Theta}_{\mathrm{ES}}$, combining structured low-rank updates with boundary-cliff retention.
\end{itemize}

\textbf{Mode II (Procedural Reservoir + Compact Ridge Head).} A single $24\text{-byte}$ seed $\Theta$ synthesizes $(\mathbf{W},\mathbf{b})$ once via Eq.~\eqref{eq:lecun}. Letting $\mathbf{H} = \tanh(\mathbf{X}_{\mathrm{tr}}\mathbf{W} + \mathbf{1}_{N_{\mathrm{tr}}}\mathbf{b}^T)$ and $\tilde{\mathbf{H}} = [\mathbf{H}, \mathbf{1}_{N_{\mathrm{tr}}}] \in \mathbb{R}^{N_{\mathrm{tr}}\times(H+1)}$, we solve the closed-form ridge readout $\mathbf{W}_{\mathrm{out}} = (\tilde{\mathbf{H}}^T\tilde{\mathbf{H}} + \lambda\mathbf{I})^{-1}\tilde{\mathbf{H}}^T\mathbf{Y}$ (with $\mathbf{Y}\in\{-1,+1\}^{N_{\mathrm{tr}}\times 1}$ for binary or one-hot $\{0,1\}^{N_{\mathrm{tr}}\times C}$ for multi-class). Symmetric INT8 quantization stores $\mathbf{W}_q = \mathrm{clip}(\mathrm{round}(\mathbf{W}_{\mathrm{out}}/s_q), -127, 127)$ with $s_q = (\max_{i,j}|[\mathbf{W}_{\mathrm{out}}]_{i,j}|+10^{-12})/127$ ($4\text{ B}$).

\begin{algorithm}[t]
\caption{FNS Boot-Time Weight Synthesis and Dual-Mode Inference}
\label{alg:fns}
\footnotesize
\begin{tabular}{@{}p{0.96\columnwidth}@{}}
\textbf{Input:} Persistent seed $\Theta=(c_x,c_y,\ell)\in\mathbb{F}_{64}^3$ ($24\text{ B}$) [or $\boldsymbol{\Theta}\in\mathbb{F}_{64}^{P\times 3}$ for Mode~I], dimensions $(d,H)$, aperture $\mathcal{A}$, cutoff $M_{\max}$, bailout $R=256$. \\
\textbf{Output:} Volatile SRAM buffer $(\mathbf{W},\mathbf{b})$ and prediction $\hat{y}$. \\
\midrule
\textbf{Phase A: One-Time Boot Synthesis (Flash $\to$ SRAM)} \\
1:\; Set $s \leftarrow 10^{-\ell}$ and $M \leftarrow (d+1)H$. \\
2:\; Compute $\{C_k\}_{k=0}^{M-1}$ via aperture $\mathcal{A}$ (Eqs.~\eqref{eq:rot}--\eqref{eq:tripod}). \\
3:\; Evaluate smooth escape potential $u_k \leftarrow \tilde{\nu}(C_k)$ (Eqs.~\eqref{eq:quad_map}--\eqref{eq:smooth_nu}). \\
4:\; Standardize and reshape $(\mathbf{W},\mathbf{b})$ via Eq.~\eqref{eq:lecun} into SRAM ($0$ extra B). \\
\textbf{Phase B: Offline Backprop-Free Training} \\
5:\; \textbf{if} Mode~I ($0\text{ B}$ readout) \textbf{then} optimize $\boldsymbol{\Theta}$ via Elitist Walk or Hybrid Rank-1 ES. \\
6:\; \textbf{else} (Mode~II) fit ridge head $\mathbf{W}_{\mathrm{out}} \leftarrow (\tilde{\mathbf{H}}^T\tilde{\mathbf{H}} + \lambda\mathbf{I})^{-1}\tilde{\mathbf{H}}^T\mathbf{Y}$; optionally quantize $\mathbf{W}_{\mathrm{out}}$ to symmetric INT8. \\
\textbf{Phase C: Runtime SRAM Inference ($O(dH)$ dot-product, $0$ fractal ops)} \\
7:\; Compute $\tilde{\mathbf{h}} = [\tanh(\mathbf{W}^T\mathbf{x} + \mathbf{b});\, 1] \in \mathbb{R}^{H+1}$ and score $\mathbf{s} = \mathbf{W}_{\mathrm{out}}^T \tilde{\mathbf{h}}$. \\
8:\; Return $\hat{y} = \mathbb{I}[s_1 \ge 0]$ if $C_{\mathrm{out}}=1$, else $\hat{y} = \arg\max_c s_c$. \\
\end{tabular}
\end{algorithm}

\section{Empirical Evaluation and Controlled Ablations}
\label{sec:experiments}

\subsection{Experimental Protocol, Hyperparameters, and Statistical Rigor}
\label{sec:protocol}

All experiments were executed deterministically in Python (NumPy~2.4.6, SciPy~1.17.1, \texttt{scikit-learn}~1.9.1 \cite{pedregosa2011scikit}) on an Intel64 host (Family~6 Model~78, 4 logical CPUs, Windows~10). All continuous and real-world evaluations employ strict, leak-free 10-fold stratified cross-validation (unified across both tables and Pareto curves). Crucially, all four real-world benchmarks in Experiments~3 and~4 use native or fixed-constant input scaling requiring \textbf{0 persistent scaler bytes} in Flash. We report Mean $\pm$ SD test accuracy, Macro-F1, exact Hidden/Readout/Total Flash bytes (including $+8\text{ B}$ per quantized MLP layer for separate weight and bias FP32 scales), and uncorrected two-sided paired Wilcoxon signed-rank $p$-values \cite{demsar2006statistical} across the 10 folds (noting significance under Holm--Bonferroni family-wise correction where relevant). Fixed Mode~II hyperparameters are disclosed explicitly: anchor $\Theta_0 = (-0.743643887, 0.131825904, 2.15)$, $M_{\max}=70$, $R=256.0$; Experiment~1 uses $H=64$, $\gamma=2.8$, $\lambda=10^{-3}$, RFF scale $1.8$, converged MLP (\texttt{ReLU}, \texttt{L-BFGS}, $\alpha=10^{-3}$, \texttt{max\_iter}=1000); Experiments~3--4 use $H=256$, $\gamma=1.8$, $\lambda=0.5$, RFF scale $1.4/\sqrt{d}$, converged MLP (\texttt{ReLU}, Adam \cite{kingma2015adam}, $\alpha=10^{-3}$, \texttt{max\_iter}=400).

\subsection{Experiment 1: Zero-Readout Boolean Logic and Converged-Baseline 2D Manifolds}
\label{sec:exp_1}

\textbf{Fully Procedural Boolean Logic (0-Byte Readout).} Across $N_{\mathrm{runs}}=10$ independent evolutionary seeds ($N_{\mathrm{pop}}=36$, verified at $128\times 128$ resolution on \textit{every} solved run), $24\text{-byte}$ 4-Quadrant synthesis solved OR ($10/10$ runs, median $3.0$ gens, IQR $1.0$), AND ($8/10$, median $13.0$, IQR $11.8$), NAND ($10/10$, median $10.5$, IQR $24.0$), and NOR ($10/10$, median $2.0$, IQR $1.8$). Composing three synthesized $24\text{-byte}$ seeds (OR and NAND feeding a synthesized AND output neuron) with pure continuous sigmoid activations (fixed gain $3.0$, no hard binarization) resolves the non-linear XOR barrier \cite{minsky1969perceptrons} at \textbf{$100.0\%$ accuracy in $72\text{ B}$ total storage and $0\text{ B}$ readout}, yielding output probabilities $[0.020, 0.926, 0.932, 0.145]$ (minimum decision margin $0.355$).

\textbf{2D Manifolds Against Converged Baselines.} Table~\ref{tab:manifolds} and Figure~\ref{fig:manifolds}(a)--(b) report 10-fold CV results on Two-Moons ($N=1000, \sigma=0.10$) and Two-Spirals \cite{lang1988learning} ($N=200, \sigma=0.04$) with $H=64$. On Two-Moons, FNS-Grid (FP32) achieves $99.80\%\pm 0.63\%$ ($284\text{ B}$, $p=1.000$ vs.\ converged MLP) and FNS-Spiral (INT8) reaches \textbf{$99.00\%\pm 1.25\%$ in $93\text{ B}$} ($+11.60$ pp over $0\text{-hidden}$ Linear-INT8 at $87.40\%$, while matching converged FP32 MLP's $99.90\%$ at $11.1\times$ smaller Flash). On Two-Spirals, FNS-Grid (FP32) achieves $96.00\%\pm 5.16\%$ ($+30.00$ pp over Linear-INT8's $66.00\%$, $p=0.0625$ vs.\ converged MLP's $99.50\%$). Equally important, two honest limitations appear in Table~\ref{tab:manifolds}: (i)~aggressive INT8 readout quantization on Two-Spirals degrades FNS-Spiral sharply ($91.00\% \to 58.50\%$), showing high-curvature winding boundaries require FP32 readouts, and (ii)~with fixed unoptimized hidden layers, $8\text{-byte}$ Seeded-PRNG RFF matches converged MLP ($99.90\%$ / $99.50\%$) in $77\text{ B}$.

\begin{table}[t]
\centering
\caption{Experiment 1: Strict 10-Fold Stratified CV on 2D Manifolds ($H=64$, Converged MLP Baseline, $+8\text{ B}$/Layer Quantized MLP Scales).}
\label{tab:manifolds}
\resizebox{\columnwidth}{!}{
\begin{tabular}{@{}llcccc@{}}
\toprule
\textbf{Dataset} & \textbf{Model} & \textbf{Acc.\ (\%)} & \textbf{F1 (\%)} & \textbf{H / R / T (B)} & \textbf{$p$ vs.\ MLP} \\
\midrule
\multirow{7}{*}{Two-Moons} & FNS-Grid (FP32) & $99.80\pm 0.63$ & 99.80 & 24 / 260 / 284 & 1.0000 \\
 & FNS-Spiral (INT8) & $99.00\pm 1.25$ & 99.00 & 24 / 69 / \textbf{93} & 0.0312 \\
 & Linear-INT8 (0-Hid) & $87.40\pm 3.41$ & 87.38 & 0 / 7 / 7 & 0.0020 \\
 & Seeded-RFF (INT8) & $99.90\pm 0.32$ & 99.90 & 8 / 69 / 77 & 1.0000 \\
 & MLP (FP32, L-BFGS) & $99.90\pm 0.32$ & 99.90 & 768 / 260 / 1,028 & --- \\
 & MLP (INT8) & $99.90\pm 0.32$ & 99.90 & 200 / 73 / 273 & 1.0000 \\
 & MLP (INT4) & $99.60\pm 0.97$ & 99.60 & 104 / 41 / 145 & 0.5000 \\
\midrule
\multirow{7}{*}{Two-Spirals} & FNS-Grid (FP32) & $96.00\pm 5.16$ & 95.98 & 24 / 260 / 284 & 0.0625 \\
 & FNS-Spiral (FP32) & $91.00\pm 6.58$ & 90.95 & 24 / 260 / 284 & 0.0078 \\
 & FNS-Spiral (INT8) & $58.50\pm 14.54$ & 57.30 & 24 / 69 / 93 & 0.0020 \\
 & Linear-INT8 (0-Hid) & $66.00\pm 7.38$ & 65.63 & 0 / 7 / 7 & 0.0020 \\
 & Seeded-RFF (INT8) & $99.50\pm 1.58$ & 99.50 & 8 / 69 / 77 & 1.0000 \\
 & MLP (FP32, L-BFGS) & $99.50\pm 1.58$ & 99.50 & 768 / 260 / 1,028 & --- \\
 & MLP (INT4) & $95.50\pm 5.50$ & 95.45 & 104 / 41 / 145 & 0.0625 \\
\bottomrule
\end{tabular}
}
\end{table}

\subsection{Experiment 2: Generator Ablation ($\partial\mathcal{M}$ vs.\ PRNG, Fourier, and Lattice Noise)}
\label{sec:exp_gen}

A central baseline question is what distinguishes $\partial\mathcal{M}$ from other $24\text{-byte}$ deterministic generators: a SHA-256 PRNG keyed by $\Theta$, a fixed smooth 2D Fourier basis (6 harmonics, 0 LUT), and a fixed $64\times 64$ $C^1$ smoothstep value-noise lattice ($16\text{ KB}$ FP32 LUT). We evaluated all four across $\ell \in \{0,\dots,8\}$ around $c_{\mathrm{Mis}}$ ($M_{\max}(\zeta)=70+40\ell$, 12 trials/scale), measuring: (a)~the standardized $16\times 16$ effective matrix rank after applying Eq.~\eqref{eq:lecun}, (b)~local coordinate correlation $\rho(\Theta,\Theta+\boldsymbol{\delta}) = \mathrm{corr}(\mathbf{u}(\Theta), \mathbf{u}(\Theta+\boldsymbol{\delta}))$ for micro-step $\boldsymbol{\delta}=(0.02s,0,0)$, and (c)~a $48\text{-byte}$ zero-readout continuous domain-drift tracking task applying \textbf{identical zero-byte z-score standardization across all four generators} on Two-Moons rotated through $\theta \in \{0^\circ, 15^\circ, 30^\circ, 45^\circ, 60^\circ\}$ ($N_{\mathrm{runs}}=10$, 10 online steps/drift, plus $1\%$ micro-step perturbation retention from each discovered $\theta=0^\circ$ anchor).

\begin{figure*}[t]
\centering
\includegraphics[width=\textwidth]{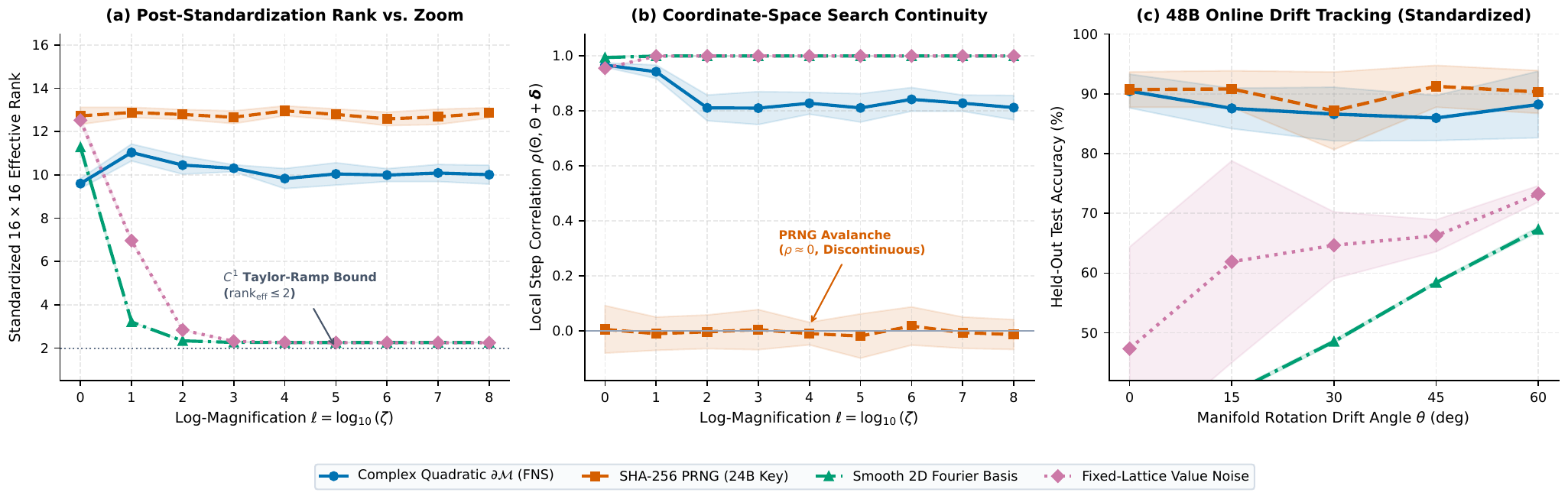}
\caption{Experiment 2 Generator Ablation ($24\text{-byte}$ seed budget, identical z-score standardization). (a)~Post-standardization $16\times 16$ effective rank across $\ell \in [0,8]$: smooth Fourier and fixed-lattice noise collapse onto the $C^1$ Taylor-ramp bound ($\mathrm{rank}_{\mathrm{eff}}\approx 2.25$), while $\partial\mathcal{M}$ holds steady at $\approx 10.0$. (b)~Local step correlation $\rho(\Theta,\Theta+\boldsymbol{\delta})$: SHA-256 PRNG exhibits avalanche discontinuity ($\rho\approx 0$), whereas $\partial\mathcal{M}$ preserves smooth navigability ($\rho\in[0.81, 0.97]$). (c)~48B zero-readout online drift tracking across rotated Two-Moons manifolds ($N_{\mathrm{runs}}=10$, Mean $\pm$ SD).}
\label{fig:gen_ablation}
\end{figure*}

Figure~\ref{fig:gen_ablation} and Table~\ref{tab:gen_ablation} establish three sharp structural separations:
\begin{enumerate}
    \item \textbf{Post-Standardization $C^1$ Taylor-Ramp Collapse:} For any $C^1$ generator $g$, zooming to $s\to 0$ yields $g(c_0 + s\mathbf{p}_k) = g(c_0) + s\nabla g(c_0)^T\mathbf{p}_k + O(s^2)$. Even \textit{after} dividing by $\sigma(\mathbf{u})$ via Eq.~\eqref{eq:lecun}, standardized patches degenerate into a planar ramp $\hat{\mathbf{g}}^T\mathbf{p}_k$ of matrix rank $\le 2$. Empirically, both Smooth Fourier ($\mathrm{rank}_{\mathrm{eff}} = 2.26$) and Fixed-Lattice Value Noise ($2.25$, bounded by finite spatial bandwidth and $C^1$ smoothstep interpolation) collapse onto this bound for $\ell \ge 3$, degrading 48B drift tracking to $53.54\%$ and $66.51\%$, respectively.
    \item \textbf{Scale-Invariant Unnormalized Variance of $\partial\mathcal{M}$:} While normalized variance $\mathrm{Var}(\tilde{\nu})$ decreases gently as $\Theta((\log\zeta)^{-2})$ due to dividing by $M_{\max}(\zeta)$ ($1.82\times 10^{-3}$ at $\ell=6$, $>2.5\times 10^8\times$ above lattice noise and $>1.3\times 10^9\times$ above Fourier), the unnormalized escape-time variance $M_{\max}(\zeta)^2\mathrm{Var}(\tilde{\nu})$ remains constant within $[173.6, 194.1]$ across six full decades ($\ell\in\{2,\dots,8\}$), sustaining post-standardization effective rank $9.99\text{--}10.01$.
    \item \textbf{PRNG Avalanche vs.\ Smooth Perturbation Retention:} SHA-256 PRNG maintains high rank ($12.58$) but exhibits zero local continuity ($\rho = 0.019\pm 0.069$). Perturbing a discovered 48B anchor by a $1\%$ coordinate step causes PRNG accuracy to collapse instantaneously to \textbf{$50.60\%\pm 20.18\%$ (chance)}, whereas FNS retains \textbf{$89.15\%\pm 3.06\%$ ($+38.55$ pp stability advantage)}. Conversely, we report transparently that on this tiny 33-parameter 2D problem, 120 blind random draws per drift step still allow PRNG random shooting to reach $89.88\%\pm 3.00\%$ versus FNS's $87.07\%\pm 2.46\%$. Thus $\partial\mathcal{M}$'s unique advantage is joint multi-scale rank \textit{and} local manifold continuity with zero lookup tables.
\end{enumerate}

\begin{table}[t]
\centering
\caption{Experiment 2: Multi-Scale Generator Properties ($\ell=6$, $\zeta=10^6$) and 48B Standardized Zero-Readout Tracking ($N_{\mathrm{runs}}=10$).}
\label{tab:gen_ablation}
\resizebox{\columnwidth}{!}{
\begin{tabular}{@{}lccccc@{}}
\toprule
\textbf{Generator (24B)} & \textbf{Rank$_{16}$ ($\ell{=}6$)} & \textbf{$\rho(\boldsymbol{\delta})$ ($\ell{=}6$)} & \textbf{1\%-Step Ret.} & \textbf{Drift Acc.} & \textbf{LUT} \\
\midrule
\textbf{Quadratic $\partial\mathcal{M}$} & $\mathbf{9.99\pm 0.30}$ & $\mathbf{0.842\pm 0.043}$ & $\mathbf{89.15\pm 3.06\%}$ & $87.07\pm 2.46\%$ & \textbf{0 B} \\
SHA-256 PRNG & $12.58\pm 0.31$ & $0.019\pm 0.069$ & $50.60\pm 20.18\%$ & $89.88\pm 3.00\%$ & 0 B \\
Smooth Fourier & $2.26\pm 0.00$ & $1.000\pm 0.000$ & $29.25\pm 0.35\%$ & $53.54\pm 0.20\%$ & 0 B \\
Lattice Noise & $2.25\pm 0.00$ & $1.000\pm 0.000$ & $47.35\pm 17.01\%$ & $66.51\pm 5.41\%$ & 16 KB \\
\bottomrule
\end{tabular}
}
\end{table}

\subsection{Experiment 3: Four Geometric Apertures and Spectral Conditioning}
\label{sec:exp_apertures}

\begin{figure*}[t]
\centering
\includegraphics[width=\textwidth]{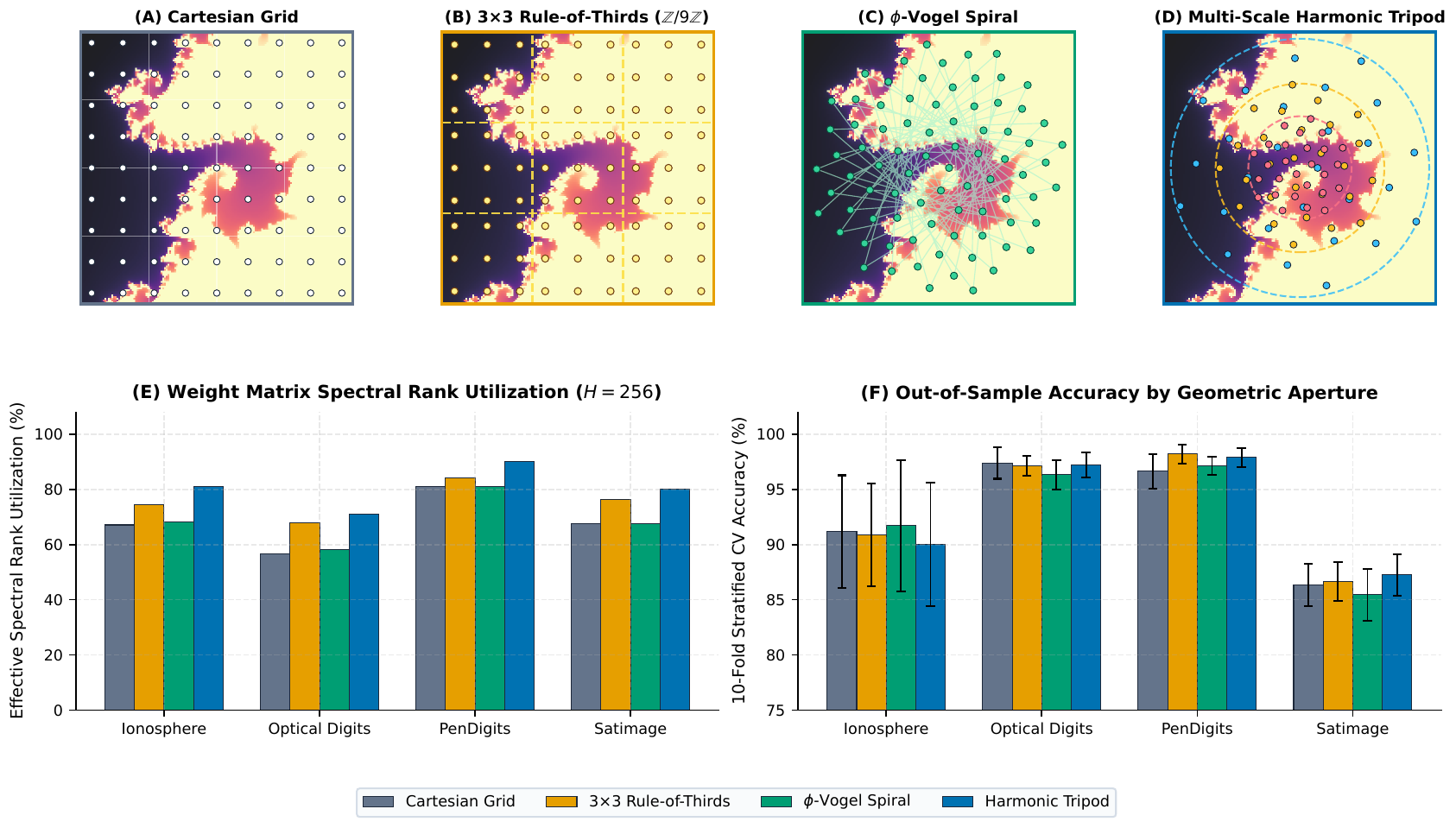}
\caption{Experiment 3 Geometric Apertures Ablation ($H=256$, identical $24\text{ B}$ seed). Top (A--D): Query stencils overlaid on the Seahorse Valley escape-time field ($R=256$). Bottom: (E)~Effective spectral rank utilization and (F)~10-fold stratified CV test accuracy across four zero-scaler-byte real-world benchmarks.}
\label{fig:apertures}
\end{figure*}

We evaluated all four sampling apertures at $H=256$ using the identical fixed $24\text{-byte}$ seed across four zero-scaler-byte real-world datasets (Figure~\ref{fig:apertures} and Table~\ref{tab:apertures}), computing effective rank $\mathrm{rank}_{\mathrm{eff}}(\mathbf{W}) = \exp(-\sum_{j=1}^r p_j \ln p_j)$ \cite{roy2007effective}, mean absolute off-diagonal column correlation, and paired Wilcoxon $p$-values versus the Cartesian Grid baseline.

Two consistent empirical patterns emerge:
\begin{itemize}
    \item \textbf{Harmonic Tripod Maximizes Spectral Conditioning Across All Four Datasets:} Interleaving three concentric zoom tiers achieves the highest effective rank utilization ($81.0\%$ Ionosphere, $70.9\%$ Optical Digits, $90.1\%$ PenDigits, $80.0\%$ Satimage---up to $+25.5\%$ relative rank gain over Cartesian Grid on Optical Digits) and the lowest mean absolute column correlation ($0.221\text{--}0.260$ vs.\ $0.316\text{--}0.340$ for Cartesian Grid and $0.351\text{--}0.354$ for single-scale $\phi$-Vogel Spiral, whose slightly higher correlation reflects Fibonacci-lag parastichy alignment).
    \item \textbf{Statistically Significant Accuracy Gains on Multi-Class Trajectories:} On UCI PenDigits ($d=16, C=10$), both coprime $3\times 3$ Rule-of-Thirds (\textbf{$98.20\%\pm 0.89\%$, $p=0.0117$}) and Harmonic Tripod (\textbf{$97.90\%\pm 0.84\%$, $p=0.0234$}) significantly outperform Cartesian Grid ($96.65\%\pm 1.56\%$). On Statlog Satimage ($C=6$), Harmonic Tripod achieves the top accuracy ($87.25\%\pm 1.89\%$, $p=0.1797$), while all four apertures are statistically comparable on Ionosphere and Optical Digits ($p \ge 0.0625$).
\end{itemize}

\begin{table}[t]
\centering
\caption{Experiment 3: Spectral Conditioning and 10-Fold Stratified CV Accuracy Across Four Apertures ($H=256$, Identical 24B Seed).}
\label{tab:apertures}
\resizebox{\columnwidth}{!}{
\begin{tabular}{@{}llccccc@{}}
\toprule
\textbf{Dataset ($d$)} & \textbf{Aperture} & \textbf{Acc.\ (\%)} & \textbf{Eff.\ Rank} & \textbf{Rank Util.} & \textbf{$|\mathrm{Corr}|$} & \textbf{$p$ vs.\ Grid} \\
\midrule
\multirow{4}{*}{Ionosphere ($34$)} & Cartesian Grid & $91.17\pm 5.10$ & 22.82 & 67.1\% & 0.324 & --- \\
 & 3$\times$3 Rule-of-Thirds & $90.87\pm 4.64$ & 25.29 & 74.4\% & 0.422 & 1.0000 \\
 & $\phi$-Vogel Spiral & $\mathbf{91.71\pm 5.94}$ & 23.19 & 68.2\% & 0.353 & 0.9590 \\
 & Harmonic Tripod & $90.02\pm 5.61$ & \textbf{27.55} & \textbf{81.0\%} & \textbf{0.226} & 0.3438 \\
\midrule
\multirow{4}{*}{Opt.\ Digits ($64$)} & Cartesian Grid & $\mathbf{97.38\pm 1.42}$ & 36.15 & 56.5\% & 0.340 & --- \\
 & 3$\times$3 Rule-of-Thirds & $97.16\pm 0.89$ & 43.48 & 67.9\% & 0.421 & 0.6445 \\
 & $\phi$-Vogel Spiral & $96.33\pm 1.34$ & 37.23 & 58.2\% & 0.351 & 0.0625 \\
 & Harmonic Tripod & $97.22\pm 1.11$ & \textbf{45.37} & \textbf{70.9\%} & \textbf{0.221} & 0.6406 \\
\midrule
\multirow{4}{*}{PenDigits ($16$)} & Cartesian Grid & $96.65\pm 1.56$ & 12.98 & 81.1\% & 0.328 & --- \\
 & 3$\times$3 Rule-of-Thirds & $\mathbf{98.20\pm 0.89}$ & 13.46 & 84.1\% & 0.415 & \textbf{0.0117} \\
 & $\phi$-Vogel Spiral & $97.15\pm 0.82$ & 12.97 & 81.1\% & 0.354 & 0.5273 \\
 & Harmonic Tripod & $97.90\pm 0.84$ & \textbf{14.42} & \textbf{90.1\%} & \textbf{0.260} & \textbf{0.0234} \\
\midrule
\multirow{4}{*}{Satimage ($36$)} & Cartesian Grid & $86.35\pm 1.90$ & 24.28 & 67.5\% & 0.316 & --- \\
 & 3$\times$3 Rule-of-Thirds & $86.65\pm 1.75$ & 27.46 & 76.3\% & 0.416 & 0.6992 \\
 & $\phi$-Vogel Spiral & $85.45\pm 2.33$ & 24.34 & 67.6\% & 0.353 & 0.4766 \\
 & Harmonic Tripod & $\mathbf{87.25\pm 1.89}$ & \textbf{28.79} & \textbf{80.0\%} & \textbf{0.226} & 0.1797 \\
\bottomrule
\end{tabular}
}
\end{table}

\subsection{Experiment 4: Zero-Scaler Real-World Benchmarks and Fair Baselines}
\label{sec:exp_real}

\begin{figure*}[t]
\centering
\includegraphics[width=0.96\textwidth]{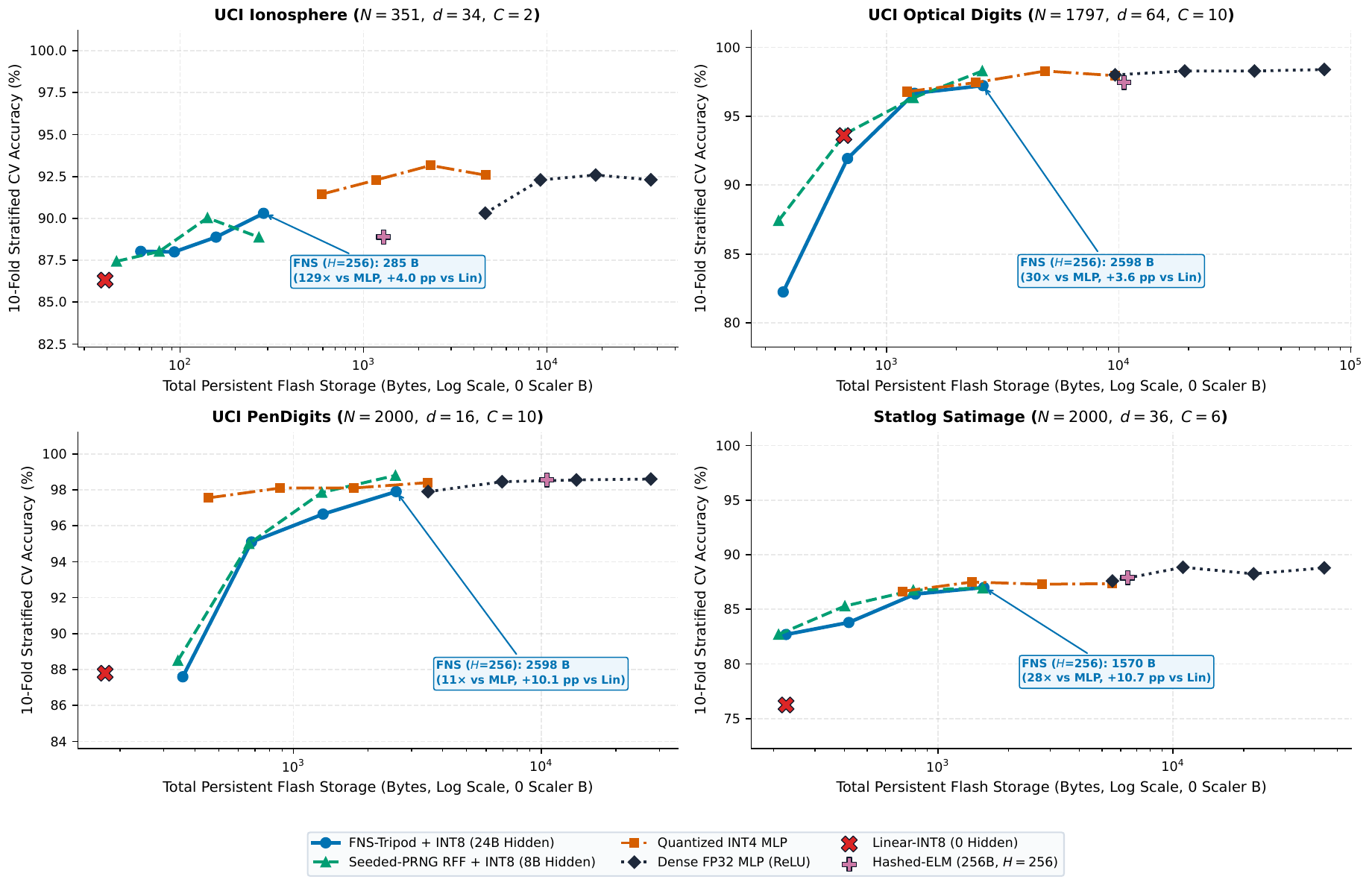}
\caption{Experiment 4 Unified 10-Fold Stratified CV Memory--Accuracy Pareto Frontiers ($H\in\{32,64,128,256\}$, $0\text{ B}$ scaler overhead) across four non-linear real-world benchmarks. FNS-Tripod + INT8 substantially outperforms Linear-INT8 ($+3.6$ to $+10.8$ pp) and achieves $10.7\times$--$129\times$ total Flash reduction ($725\times$--$2{,}773\times$ hidden reduction) relative to converged Dense FP32 MLPs.}
\label{fig:pareto}
\end{figure*}

Locking the uniform Harmonic Tripod aperture ($H=256$, $24\text{ B}$ hidden seed) without per-dataset selection, we evaluated four genuinely non-linear, zero-scaler-byte real-world benchmarks: UCI Ionosphere Radar ($N=351, d=34, C=2$, native $[-1,1]$), UCI Optical Digits ($N=1797, d=64, C=10$, fixed $X/8-1$), UCI PenDigits ($N=2000$ stratified from $10{,}992$, $d=16, C=10$, fixed $X/50-1$), and Statlog Landsat Satimage ($N=2000$ stratified from $6{,}430$, $d=36, C=6$, pre-normalized multispectral bands). We compare against $0\text{-hidden}$ Linear-INT8, $8\text{-byte}$ Seeded-PRNG RFF (INT8), $8\text{-byte}$ Seeded-PRNG ELM (INT8), Explicit RFF (FP32), fixed 2D SplitMix64 + sign-hash Untrained Hashed-ELM ($256\text{ B}$ hidden), and converged FP32/INT8/INT4 MLPs (Table~\ref{tab:realworld} and unified 10-fold Pareto curves in Figure~\ref{fig:pareto}).

\begin{table*}[t]
\centering
\caption{Experiment 4: Strict 10-Fold Stratified CV Across Four Zero-Scaler-Byte Non-Linear Benchmarks ($H=256$). Exact Hidden / Readout / Total persistent Flash bytes are shown ($0\text{ B}$ scaler overhead). $p_{\mathrm{MLP}}$, $p_{\mathrm{Lin}}$, and $p_{\mathrm{RFF}}$ denote uncorrected two-sided paired Wilcoxon signed-rank $p$-values against Dense FP32 MLP, Linear-INT8, and Seeded-PRNG RFF (INT8).}
\label{tab:realworld}
\resizebox{\textwidth}{!}{
\begin{tabular}{@{}llcccccccc@{}}
\toprule
\textbf{Dataset} & \textbf{Model} & \textbf{Acc.\ (Mean $\pm$ SD)} & \textbf{Macro-F1} & \textbf{Hidden (B)} & \textbf{Readout (B)} & \textbf{Total (B)} & \textbf{$p_{\mathrm{MLP}}$} & \textbf{$p_{\mathrm{Lin}}$} & \textbf{$p_{\mathrm{RFF}}$} \\
\midrule
\multirow{9}{*}{\shortstack[l]{\textbf{UCI Ionosphere}\\($N=351, d=34, C=2$)\\Native $[-1,1]$ (0 B)}}
 & \textbf{FNS-Tripod (INT8)} & $90.30 \pm 5.09\%$ & 88.65\% & \textbf{24} & 261 & \textbf{285} ($129\times$) & 0.2031 & \textbf{0.0254} & 0.5312 \\
 & FNS-Tripod (FP32) & $90.02 \pm 5.61\%$ & 88.22\% & \textbf{24} & 1,028 & 1,052 & 0.2031 & 0.0430 & 0.7422 \\
 & Linear-INT8 (0-Hid) & $86.32 \pm 4.84\%$ & 83.62\% & 0 & 39 & 39 & 0.0039 & --- & 0.1875 \\
 & Seeded-RFF (INT8) & $88.87 \pm 4.38\%$ & 86.83\% & 8 & 261 & 269 & 0.0391 & 0.1875 & --- \\
 & Seeded-ELM (INT8) & $90.32 \pm 3.34\%$ & 88.79\% & 8 & 261 & 269 & 0.0938 & 0.0703 & 0.3438 \\
 & Hashed-ELM (256B, 2D) & $88.89 \pm 3.91\%$ & 86.89\% & 256 & 1,028 & 1,284 & 0.0625 & 0.1797 & 0.9922 \\
 & Dense MLP (FP32) & $92.30 \pm 3.83\%$ & 91.24\% & 35,840 & 1,028 & 36,868 & --- & 0.0039 & 0.0391 \\
 & Quantized MLP (INT8) & $92.30 \pm 3.83\%$ & 91.24\% & 8,968 & 265 & 9,233 & 1.0000 & 0.0039 & 0.0391 \\
 & Quantized MLP (INT4) & $92.58 \pm 4.10\%$ & 91.56\% & 4,488 & 137 & 4,625 & 1.0000 & 0.0039 & 0.0234 \\
\midrule
\multirow{9}{*}{\shortstack[l]{\textbf{UCI Optical Digits}\\($N=1797, d=64, C=10$)\\Fixed $X/8-1$ (0 B)}}
 & \textbf{FNS-Tripod (INT8)} & $97.22 \pm 1.11\%$ & 97.20\% & \textbf{24} & 2,574 & \textbf{2,598} ($29.6\times$) & 0.0039 & \textbf{0.0020} & 0.0078 \\
 & FNS-Tripod (FP32) & $97.22 \pm 1.11\%$ & 97.20\% & \textbf{24} & 10,280 & 10,304 & 0.0039 & 0.0020 & 0.0078 \\
 & Linear-INT8 (0-Hid) & $93.60 \pm 1.33\%$ & 93.53\% & 0 & 654 & 654 & 0.0020 & --- & 0.0020 \\
 & Seeded-RFF (INT8) & $98.27 \pm 1.00\%$ & 98.26\% & 8 & 2,574 & 2,582 & 1.0000 & 0.0020 & --- \\
 & Seeded-ELM (INT8) & $98.22 \pm 0.97\%$ & 98.21\% & 8 & 2,574 & 2,582 & 0.7773 & 0.0020 & 0.5938 \\
 & Hashed-ELM (256B, 2D) & $97.44 \pm 0.95\%$ & 97.44\% & 256 & 10,280 & 10,536 & 0.0078 & 0.0020 & 0.0117 \\
 & Dense MLP (FP32) & $98.39 \pm 0.81\%$ & 98.38\% & 66,560 & 10,280 & 76,840 & --- & 0.0020 & 1.0000 \\
 & Quantized MLP (INT8) & $98.33 \pm 0.87\%$ & 98.32\% & 16,648 & 2,578 & 19,226 & 1.0000 & 0.0020 & 0.8750 \\
 & Quantized MLP (INT4) & $97.94 \pm 0.53\%$ & 97.93\% & 8,328 & 1,293 & 9,621 & 0.0391 & 0.0020 & 0.2969 \\
\midrule
\multirow{9}{*}{\shortstack[l]{\textbf{UCI PenDigits}\\($N=2000, d=16, C=10$)\\Fixed $X/50-1$ (0 B)}}
 & \textbf{FNS-Tripod (INT8)} & $97.90 \pm 0.97\%$ & 97.89\% & \textbf{24} & 2,574 & \textbf{2,598} ($10.7\times$) & 0.1094 & \textbf{0.0020} & 0.0078 \\
 & FNS-Tripod (FP32) & $97.90 \pm 0.84\%$ & 97.90\% & \textbf{24} & 10,280 & 10,304 & 0.0469 & 0.0020 & 0.0039 \\
 & Linear-INT8 (0-Hid) & $87.80 \pm 2.46\%$ & 87.53\% & 0 & 174 & 174 & 0.0020 & --- & 0.0020 \\
 & Seeded-RFF (INT8) & $98.80 \pm 0.95\%$ & 98.80\% & 8 & 2,574 & 2,582 & 0.5938 & 0.0020 & --- \\
 & Seeded-ELM (INT8) & $98.50 \pm 0.75\%$ & 98.50\% & 8 & 2,574 & 2,582 & 0.6406 & 0.0020 & 0.1562 \\
 & Hashed-ELM (256B, 2D) & $98.55 \pm 0.93\%$ & 98.54\% & 256 & 10,280 & 10,536 & 1.0000 & 0.0020 & 0.4062 \\
 & Dense MLP (FP32) & $98.60 \pm 0.74\%$ & 98.58\% & 17,408 & 10,280 & 27,688 & --- & 0.0020 & 0.5938 \\
 & Quantized MLP (INT8) & $98.60 \pm 0.74\%$ & 98.58\% & 4,360 & 2,578 & 6,938 & 1.0000 & 0.0020 & 0.5938 \\
 & Quantized MLP (INT4) & $98.40 \pm 0.57\%$ & 98.38\% & 2,184 & 1,293 & 3,477 & 0.2188 & 0.0020 & 0.1875 \\
\midrule
\multirow{9}{*}{\shortstack[l]{\textbf{Statlog Satimage}\\($N=2000, d=36, C=6$)\\Pre-norm.\ bands (0 B)}}
 & \textbf{FNS-Tripod (INT8)} & $87.00 \pm 1.94\%$ & 83.21\% & \textbf{24} & 1,546 & \textbf{1,570} ($28.1\times$) & 0.0215 & \textbf{0.0020} & 0.9844 \\
 & FNS-Tripod (FP32) & $87.25 \pm 1.89\%$ & 83.46\% & \textbf{24} & 6,168 & 6,192 & 0.0586 & 0.0020 & 0.8691 \\
 & Linear-INT8 (0-Hid) & $76.25 \pm 2.04\%$ & 60.54\% & 0 & 226 & 226 & 0.0020 & --- & 0.0020 \\
 & Seeded-RFF (INT8) & $86.95 \pm 1.91\%$ & 83.11\% & 8 & 1,546 & 1,554 & 0.0391 & 0.0020 & --- \\
 & Seeded-ELM (INT8) & $88.05 \pm 1.76\%$ & 84.99\% & 8 & 1,546 & 1,554 & 0.3184 & 0.0020 & 0.0762 \\
 & Hashed-ELM (256B, 2D) & $87.90 \pm 2.33\%$ & 84.59\% & 256 & 6,168 & 6,424 & 0.1699 & 0.0020 & 0.5547 \\
 & Dense MLP (FP32) & $88.80 \pm 2.21\%$ & 86.56\% & 37,888 & 6,168 & 44,056 & --- & 0.0020 & 0.0391 \\
 & Quantized MLP (INT8) & $88.75 \pm 2.36\%$ & 86.51\% & 9,480 & 1,550 & 11,030 & 1.0000 & 0.0020 & 0.0430 \\
 & Quantized MLP (INT4) & $87.35 \pm 3.22\%$ & 84.89\% & 4,744 & 779 & 5,523 & 0.1406 & 0.0020 & 0.7656 \\
\bottomrule
\end{tabular}
}
\end{table*}

Three clear empirical findings emerge from Table~\ref{tab:realworld} and Figure~\ref{fig:pareto}:
\begin{enumerate}
    \item \textbf{Substantial Non-Linear Gains Over Linear-INT8:} Across every benchmark, FNS-Tripod (INT8) statistically significantly outperforms $0\text{-hidden}$ Linear-INT8: $+3.98$ pp on Ionosphere ($90.30\%$ vs.\ $86.32\%$, $p=0.0254$), $+3.62$ pp on Optical Digits ($97.22\%$ vs.\ $93.60\%$, $p=0.0020$), $+10.10$ pp on PenDigits ($97.90\%$ vs.\ $87.80\%$, $p=0.0020$), and $+10.75$ pp on Satimage ($87.00\%$ vs.\ $76.25\%$, $p=0.0020$---surviving Holm--Bonferroni correction on all three multi-class tasks).
    \item \textbf{Near-MLP Accuracy at $10.7\times$--$129\times$ Total Compression:} FNS-Tripod (INT8) stays within $0.70\text{--}2.00$ pp of converged Dense FP32 MLP while reducing hidden Flash by $725\times$--$2{,}773\times$ ($24\text{ B}$ vs.\ $17.4\text{--}66.6\text{ KB}$) and total Flash by $10.7\times$ (PenDigits, $p=0.1094$, statistically indistinguishable), $28.1\times$ (Satimage), $29.6\times$ (Optical Digits), and $129\times$ (Ionosphere, $p=0.2031$, indistinguishable). Compared against INT4 MLPs, FNS-Tripod (INT8) remains $1.34\times$--$16.2\times$ smaller while incurring virtually zero INT8 readout degradation ($0.00$ pp on Optical Digits and PenDigits; $+0.28$ pp on Ionosphere).
    \item \textbf{Candid Comparison Against Seeded-PRNG RFF/ELM and 2D Hashed-ELM:} Including fair procedural baselines reveals an important systems nuance. On Ionosphere ($90.30\%$ vs.\ $88.87\%$, $p=0.5312$) and Satimage ($87.00\%$ vs.\ $86.95\%$, $p=0.9844$), FNS-Tripod (INT8) edges or matches Seeded-PRNG RFF, whereas on Optical Digits ($-1.05$ pp, $p=0.0078$) and PenDigits ($-0.90$ pp, $p=0.0078$), $8\text{-byte}$ Seeded-PRNG RFF slightly outperforms fixed-seed FNS in $16\text{ B}$ less Flash, and properly implemented 2D SplitMix64 Hashed-ELM performs competitively ($88.89\%\text{--}98.55\%$) at $10.7\times$ larger hidden storage ($256\text{ B}$). Consequently, in fixed-seed Mode~II, counter-based PRNG reservoirs remain a strong lightweight alternative; FNS's differentiating strength lies in PRNG-free analytic determinism (immune to cross-compiler PRNG state drift) and continuous coordinate-space navigability in Mode~I (Experiment~2).
\end{enumerate}

\subsection{Experiment 5: Equal-Budget Multi-Patch Optimization Scaling}
\label{sec:exp_es}

\begin{figure*}[t]
\centering
\includegraphics[width=\textwidth]{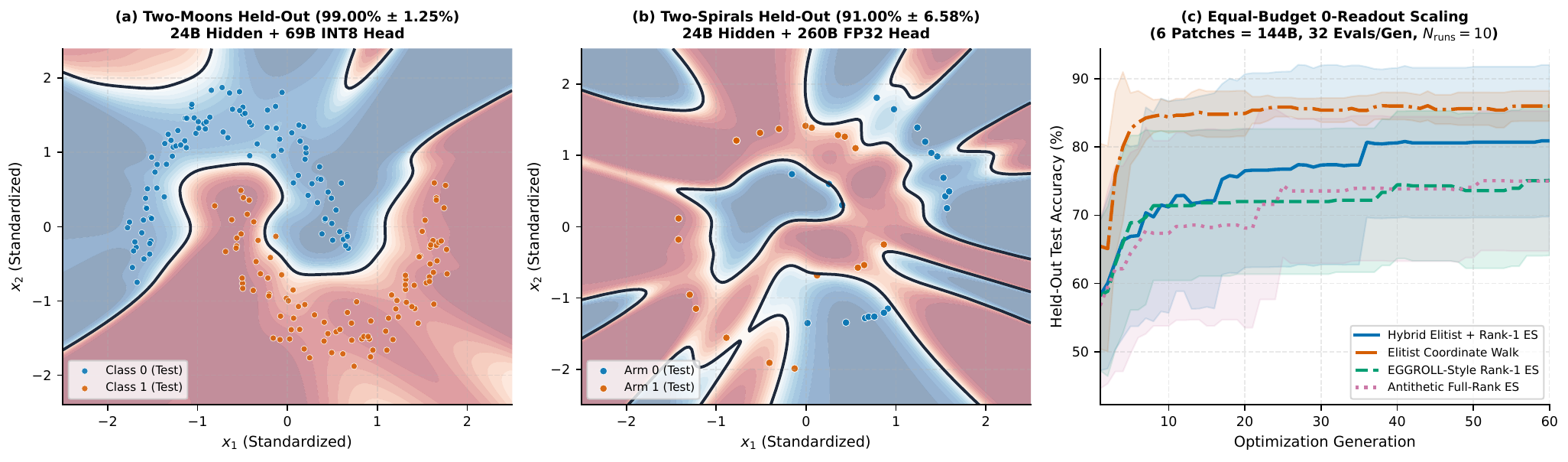}
\caption{Experiments 1 \& 5. (a)~Held-out Two-Moons contour ($24\text{ B}$ hidden + $69\text{ B}$ INT8 head, $99.00\%\pm 1.25\%$ 10-fold CV). (b)~Held-out Two-Spirals contour ($24\text{ B}$ hidden + $260\text{ B}$ FP32 head, $91.00\%\pm 6.58\%$). (c)~Strict equal-budget ($32$ evals/gen) multi-patch zero-readout scaling ($6$ patches = $144\text{ B}$, $N_{\mathrm{runs}}=10$): Hybrid Elitist + Rank-1 ES recovers $+5.80$ pp over pure Rank-1 ES while Elitist Walk yields the tightest variance.}
\label{fig:manifolds}
\end{figure*}

Evaluating a 6-patch ($18\text{D}$, $144\text{ B}$ total, $0\text{ B}$ readout) network ($2\to 40\to 1$) on stratified Two-Moons under a strict equal budget of $32$ candidate evaluations per generation ($60$ gens, $N_{\mathrm{runs}}=10$; Figure~\ref{fig:manifolds}(c)):
\begin{itemize}
    \item \textbf{Elitist Coordinate Walk} achieves \textbf{$86.00\%\pm 2.22\%$} test accuracy (BCE $0.2691$, $1{,}751\text{ ms}$), significantly outperforming Antithetic Full-Rank ES ($75.00\%\pm 10.26\%$, BCE $0.4455$, $p=0.0039$).
    \item \textbf{EGGROLL-Style Rank-1 ES} ($75.10\%\pm 11.01\%$, BCE $0.4622$, $1{,}488\text{ ms}$) is statistically tied with Antithetic Full-Rank ES ($p=0.5156$), confirming that symmetrical $\pm\boldsymbol{\epsilon}$ averaging suffers variance near steep $\partial\mathcal{M}$ boundary transitions even with large-bailout smoothing.
    \item \textbf{Hybrid Elitist + Rank-1 ES} bridges this gap cleanly, lifting test accuracy to \textbf{$80.90\%\pm 11.13\%$} and reducing BCE loss by $29\%$ ($0.4622 \to 0.3278$) at identical wall-clock speed ($1{,}502\text{ ms}$, $p=0.0801$ vs.\ Full-Rank ES, $p=0.1094$ vs.\ Elitist Walk).
\end{itemize}

\subsection{Warmed-Up Boot Synthesis and Apples-to-Apples SRAM Profiling}
\label{sec:systems}

\begin{figure*}[t]
\centering
\includegraphics[width=0.96\textwidth]{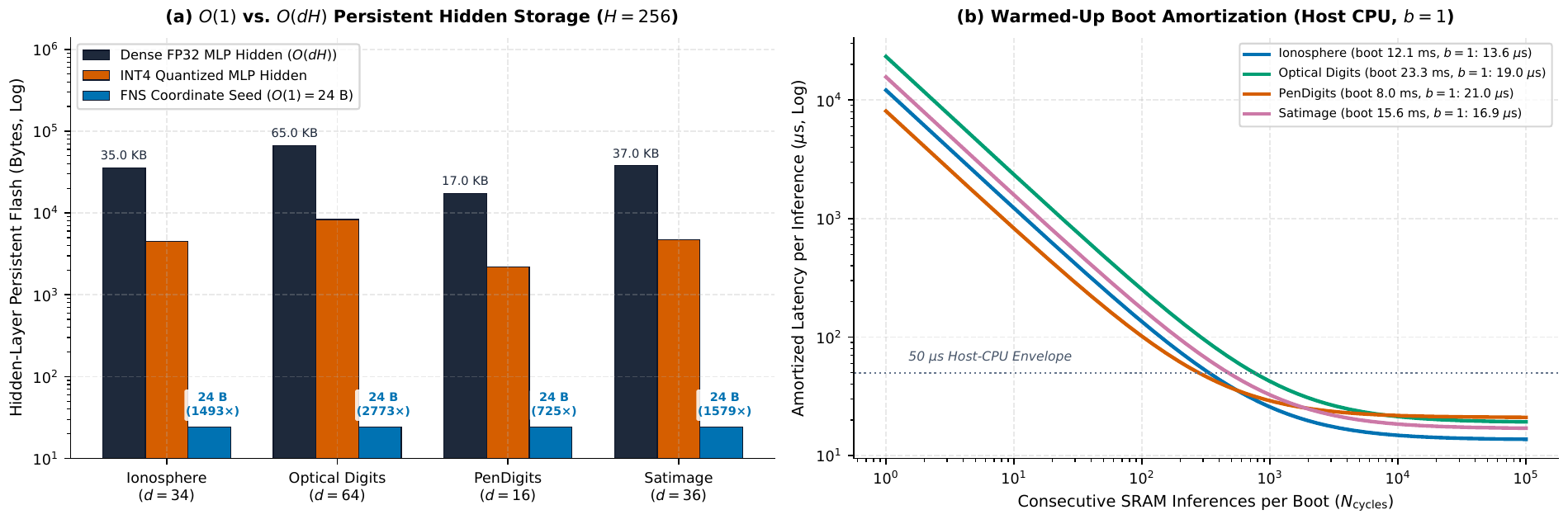}
\caption{Systems profiling from Experiment 4 ($H=256$, Intel64 host). (a)~Persistent hidden-layer Flash ($24\text{ B}$ constant FNS seed vs.\ $O(dH)$ FP32/INT4 MLPs, $725\times$--$2{,}773\times$ reduction). (b)~Warmed-up boot-time amortization $(\tau_{\mathrm{boot}}/N_{\mathrm{cycles}} + \tau_{\mathrm{SRAM},b=1})$ using apples-to-apples single-sample SRAM inference.}
\label{fig:systems}
\end{figure*}

Figure~\ref{fig:systems} reports warmed-up (10 warm-up + 50 timed runs) host-CPU profiling at $H=256$. Boot synthesis scales cleanly and monotonically with parameter count $M=(d+1)H$: $8.05\pm 1.88\text{ ms}$ on PenDigits ($d=16$), $12.11\pm 0.56\text{ ms}$ on Ionosphere ($d=34$), $15.64\pm 4.00\text{ ms}$ on Satimage ($d=36$), and $23.29\pm 3.74\text{ ms}$ on Optical Digits ($d=64$). Using the \textit{exact same} pure-NumPy single-sample ($b=1$) forward-pass timer, SRAM inference latency is virtually identical between FNS ($13.61\text{--}20.96\,\mu\text{s}$) and Dense FP32 MLP ($10.78\text{--}23.29\,\mu\text{s}$), confirming equal $O(dH)$ runtime FLOPs once materialized in volatile memory (amortizing to $5.58\text{--}8.41\,\mu\text{s}$/sample in batch mode). As shown in Figure~\ref{fig:systems}(b), the one-time boot synthesis amortizes below a $50\,\mu\text{s}$ host-CPU envelope within $N_{\mathrm{cycles}} \approx 280\text{--}750$ inferences; direct bare-metal ARM Cortex-M cycle and energy benchmarking remains important future work.

\section{Related Verification and Deployment Context}
\label{sec:ecosystem}

While this paper focuses strictly on the mathematical foundations, ablations, and ML benchmarks of FNS, specialized companion studies have investigated downstream deployment constraints of the core quadratic kernel: machine-checked Lean~4 proofs of finite-iteration escape-radius invariance ($|z_n|>2 \Rightarrow |z_{n+k}|\to\infty$, verified with \texttt{0 sorry} and standard axioms only) alongside 40-core bare-metal NUMA scaling \cite{dagli2026lean4_gauntlet}; sub-millisecond System-1 natural language triage gating ahead of System-2 language models \cite{dagli2026werr_map}; and bit-exact fixed-point EVM intra-block circuit breakers \cite{dagli2026werracle}. Together, these studies motivate FNS's emphasis on lookup-free, PRNG-free deterministic weight synthesis.

\section{Limitations, Epistemic Scope, and Future Directions}
\label{sec:limitations}

Explicitly bounding where FNS is \textit{not} optimal guides disciplined adoption:
\begin{enumerate}
    \item \textbf{Information-Theoretic Ceiling:} By Proposition~\ref{prop:capacity}, each $24\text{-byte}$ seed conveys at most $192$ bits of conditional information. FNS is suited to compact System-1 classifiers, edge controllers, and deterministic guardrails rather than replacing high-entropy frontier transformers.
    \item \textbf{Readout and Feature-Scaler Storage:} Mode~II shifts Flash scaling from $O(dH + HC)$ to $O(1 + HC)$, which dominates total Flash when $C$ is large ($2{,}574\text{ B}$ readout vs.\ $24\text{ B}$ hidden on 10-class digits at $H=256$). Furthermore, datasets lacking fixed input bounds require storing per-feature scaling statistics ($8d\text{ B}$ FP32) that dense MLPs can fold into $\mathbf{W}_1$, whereas FNS must persist them unless raw or constant-scaled inputs are used.
    \item \textbf{Mode II Parity with Seeded-PRNG Reservoirs:} When hidden seeds are unoptimized, $8\text{-byte}$ counter-based PRNG RFF/ELM reservoirs achieve slightly higher accuracy on Optical Digits and PenDigits in $16\text{ B}$ less Flash. Future Mode~II work should investigate inner-fold coordinate optimization of $\Theta^*$ along $\partial\mathcal{M}$ to test whether boundary continuity unlocks data-adaptive reservoir gains unattainable by discontinuous PRNG seeds.
    \item \textbf{Curvature-Sensitive Readout Quantization \& On-Target MCU Profiling:} Two-Spirals INT8 degradation ($91.0\% \to 58.5\%$) shows that tightly wound boundaries require FP32 or mixed-precision readouts, and host-CPU Python timings must be complemented by native C/Q16.16 fixed-point benchmarking on physical ARM Cortex-M and RISC-V microcontrollers.
\end{enumerate}

\textbf{Broader Research Outlook.} Looking beyond the compact classification benchmarks evaluated in this paper, the lookup-free boundary paradigm naturally invites four longer-horizon research directions deferred to dedicated future studies: (i)~scaling pure-integer, backpropagation-free low-rank evolution strategies \cite{sarkar2025eggroll} to larger multi-layer procedural architectures; (ii)~low-bandwidth policy synchronization in decentralized multi-agent collectives and embedded scientific instrumentation where compact $24\text{-byte}$ state descriptors can be transmitted over severely constrained channels; (iii)~passive analog optical or photonic wave-interference evaluators capable of computing boundary-integral weights without iterative digital loops; and (iv)~mitigating synthetic model collapse via non-degrading analytical parameter priors in two-tier edge-reflex-to-executive architectures.

\section{Conclusion}
\label{sec:conclusion}

We formulated and systematically evaluated \textbf{Fractal Neural Synthesis (FNS)}, a constant-storage procedural hidden-layer derivation architecture based on large-bailout ($R=256$) complex quadratic boundary dynamics and four geometric sampling apertures. Through controlled multi-scale ablations and unified 10-fold stratified cross-validation across boolean logic, 2D manifolds, and four zero-scaler-byte non-linear real-world benchmarks ($d=16\text{--}64$, $C=2\text{--}10$), we established three conclusions: (1)~among $24\text{-byte}$ lookup-free generators, $\partial\mathcal{M}$ uniquely combines scale-free post-standardization matrix rank ($\approx 10.0$ across $\zeta \in [10^0,10^8]$, vs.\ $\approx 2.25$ for $C^1$ Taylor-flattened Fourier/lattice bases) with smooth local continuity ($\rho\in[0.81, 0.97]$, retaining $89.15\%$ accuracy under $1\%$ micro-steps vs.\ $50.60\%$ for SHA-256 PRNG); (2)~coprime $3\times 3$ Rule-of-Thirds and multi-scale Harmonic Tripod apertures significantly improve spectral conditioning and multi-class generalization at zero storage cost; and (3)~pairing a $24\text{-byte}$ procedural hidden layer with an INT8 linear head outperforms $0\text{-hidden}$ Linear-INT8 by $+3.6$ to $+10.8$ percentage points while delivering $10.7\times$--$129\times$ total Flash reduction ($725\times$--$2{,}773\times$ hidden reduction) relative to converged Float32 MLPs at matching runtime SRAM latency.

\section*{Acknowledgments}
Computational infrastructure and server resources supporting the broader verification suite were provided by ITouch Systems (Cukurova University Technopark, Adana, T\"{u}rkiye). The authors gratefully acknowledge Bekirhan Da\u{g}l\i, partner at ITouch Systems, for infrastructure support and resource allocation.

\section*{Funding}
This research did not receive any specific grant from funding agencies in the public, commercial, or not-for-profit sectors; computational infrastructure was made available free of charge by ITouch Systems solely for non-commercial open scientific research.

\section*{Declaration of Competing Interests and IP Disclosure}
Volkan Da\u{g}l\i\ is a co-founder and researcher affiliated with ITouch Systems and an inventor on pending patent applications filed with the Turkish Patent and Trademark Office (T\"{U}RKPATENT applications TR~2026/016285, TR~2026/016633, and TR~2026/020266) covering procedural boundary weight synthesis and deterministic edge execution architectures. The remaining authors declare no additional competing financial interests. All benchmark scripts, deterministic seeds, and datasets reported herein are openly released for independent academic verification.

\section*{Data and Code Availability}
All benchmark scripts, deterministic random seeds, raw JSON/CSV telemetry, and vector figure generators are openly archived via CERN Zenodo (\url{https://doi.org/10.5281/zenodo.22774934}) and SSRN Abstract~7659198 (\url{https://papers.ssrn.com/abstract=7659198}). All real-world datasets (UCI Ionosphere, Optical Digits, PenDigits, and Statlog Satimage) are standard public benchmarks retrieved reproducibly via \texttt{scikit-learn} and OpenML.

\bibliographystyle{IEEEtran}
\bibliography{references}

@article{mandelbrot1980fractal,
  title={Fractal aspects of the iteration of $z \mapsto \lambda z (1-z)$ for complex $\lambda$ and $z$},
  author={Mandelbrot, Benoit B.},
  journal={Annals of the New York Academy of Sciences},
  volume={357},
  number={1},
  pages={249--259},
  year={1980},
  doi={10.1111/j.1749-6632.1980.tb22365.x}
}

@book{mandelbrot1982fractal,
  title={The Fractal Geometry of Nature},
  author={Mandelbrot, Benoit B.},
  publisher={W. H. Freeman and Company},
  address={New York},
  year={1982},
  isbn={978-0716711865}
}

@book{douady1984etude,
  title={{\'E}tude dynamique des polyn{\^o}mes complexes, Parties {I} et {II}},
  author={Douady, Adrien and Hubbard, John H.},
  series={Publications Math{\'e}matiques d'Orsay},
  volume={84-02 / 85-04},
  publisher={Universit{\'e} de Paris-Sud},
  address={Orsay},
  year={1984}
}

@book{peitgen1986beauty,
  title={The Beauty of Fractals: Images of Complex Dynamical Systems},
  author={Peitgen, Heinz-Otto and Richter, Peter H.},
  publisher={Springer-Verlag},
  address={Berlin, Heidelberg},
  year={1986},
  doi={10.1007/978-3-642-61717-1}
}

@article{tanlei1990similarity,
  title={Similarity between the {Mandelbrot} set and {Julia} sets},
  author={Tan, Lei},
  journal={Communications in Mathematical Physics},
  volume={134},
  number={3},
  pages={587--617},
  year={1990},
  doi={10.1007/BF02098448}
}

@book{peitgen1992chaos,
  title={Chaos and Fractals: New Frontiers of Science},
  author={Peitgen, Heinz-Otto and J{\"u}rgens, Hartmut and Saupe, Dietmar},
  publisher={Springer-Verlag},
  address={New York},
  year={1992},
  isbn={978-0387979038}
}

@article{shishikura1998hausdorff,
  title={The {Hausdorff} dimension of the boundary of the {Mandelbrot} set and {Julia} sets},
  author={Shishikura, Mitsuhiro},
  journal={Annals of Mathematics},
  volume={147},
  number={2},
  pages={225--267},
  year={1998},
  doi={10.2307/121009}
}

@book{milnor2006dynamics,
  title={Dynamics in One Complex Variable},
  author={Milnor, John},
  edition={3rd},
  series={Annals of Mathematics Studies},
  volume={160},
  publisher={Princeton University Press},
  address={Princeton, NJ},
  year={2006},
  doi={10.1515/9781400835539}
}

@article{hurwitz1891ueber,
  title={Ueber die angen{\"a}herte {Darstellung} der {Irrationalzahlen} durch rationale {Br{\"u}che}},
  author={Hurwitz, Adolf},
  journal={Mathematische Annalen},
  volume={39},
  number={2},
  pages={279--284},
  year={1891},
  doi={10.1007/BF01206656}
}

@article{vogel1979better,
  title={A better way to construct the sunflower head},
  author={Vogel, Helmut},
  journal={Mathematical Biosciences},
  volume={44},
  number={3--4},
  pages={179--189},
  year={1979},
  doi={10.1016/0025-5564(79)90080-4}
}

@article{shannon1948mathematical,
  title={A mathematical theory of communication},
  author={Shannon, Claude E.},
  journal={Bell System Technical Journal},
  volume={27},
  number={3},
  pages={379--423},
  year={1948},
  doi={10.1002/j.1538-7305.1948.tb01338.x}
}

@book{li2019kolmogorov,
  title={An Introduction to {Kolmogorov} Complexity and Its Applications},
  author={Li, Ming and Vit{\'a}nyi, Paul M. B.},
  edition={4th},
  publisher={Springer},
  address={Cham},
  year={2019},
  doi={10.1007/978-3-030-11298-1}
}

@book{minsky1969perceptrons,
  title={Perceptrons: An Introduction to Computational Geometry},
  author={Minsky, Marvin and Papert, Seymour A.},
  publisher={MIT Press},
  address={Cambridge, MA},
  year={1969},
  isbn={978-0262630221}
}

@article{cybenko1989approximation,
  title={Approximation by superpositions of a sigmoidal function},
  author={Cybenko, George},
  journal={Mathematics of Control, Signals, and Systems},
  volume={2},
  number={4},
  pages={303--314},
  year={1989},
  doi={10.1007/BF02551274}
}

@article{leshno1993multilayer,
  title={Multilayer feedforward networks with a nonpolynomial activation function can approximate any function},
  author={Leshno, Moshe and Lin, Vladimir Ya. and Pinkus, Allan and Schocken, Shimon},
  journal={Neural Networks},
  volume={6},
  number={6},
  pages={861--867},
  year={1993},
  doi={10.1016/S0893-6080(05)80131-5}
}

@inproceedings{lang1988learning,
  title={Learning to tell two spirals apart},
  author={Lang, Kevin J. and Witbrock, Michael J.},
  booktitle={Proceedings of the 1988 Connectionist Models Summer School},
  pages={52--59},
  year={1988}
}

@article{wolpert1997no,
  title={No free lunch theorems for optimization},
  author={Wolpert, David H. and Macready, William G.},
  journal={IEEE Transactions on Evolutionary Computation},
  volume={1},
  number={1},
  pages={67--82},
  year={1997},
  doi={10.1109/4235.585892}
}

@incollection{lecun1998efficient,
  title={Efficient {BackProp}},
  author={LeCun, Yann and Bottou, L{\'e}on and Orr, Genevieve B. and M{\"u}ller, Klaus-Robert},
  booktitle={Neural Networks: Tricks of the Trade},
  series={Lecture Notes in Computer Science},
  volume={1524},
  pages={9--50},
  publisher={Springer},
  year={1998},
  doi={10.1007/3-540-49430-8_2}
}

@article{demsar2006statistical,
  title={Statistical comparisons of classifiers over multiple data sets},
  author={Dem{\v{s}}ar, Janez},
  journal={Journal of Machine Learning Research},
  volume={7},
  number={1},
  pages={1--30},
  year={2006}
}

@inproceedings{roy2007effective,
  title={The effective rank: A measure of effective dimensionality},
  author={Roy, Olivier and Vetterli, Martin},
  booktitle={Proceedings of the 15th European Signal Processing Conference (EUSIPCO)},
  pages={606--610},
  year={2007}
}

@article{shadmehr2010error,
  title={Error correction, sensory prediction, and adaptation in motor control},
  author={Shadmehr, Reza and Smith, Maurice A. and Krakauer, John W.},
  journal={Annual Review of Neuroscience},
  volume={33},
  pages={89--108},
  year={2010},
  doi={10.1146/annurev-neuro-060909-153135}
}

@article{pedregosa2011scikit,
  title={{Scikit-learn}: Machine learning in {Python}},
  author={Pedregosa, Fabian and Varoquaux, Ga{\"e}l and Gramfort, Alexandre and Michel, Vincent and Thirion, Bertrand and Grisel, Olivier and Blondel, Mathieu and Prettenhofer, Peter and Weiss, Ron and Dubourg, Vincent and Vanderplas, Jake and Passos, Alexandre and Cournapeau, David and Brucher, Matthieu and Perrot, Matthieu and Duchesnay, {\'E}douard},
  journal={Journal of Machine Learning Research},
  volume={12},
  pages={2825--2830},
  year={2011}
}

@inproceedings{kingma2015adam,
  title={{Adam}: A method for stochastic optimization},
  author={Kingma, Diederik P. and Ba, Jimmy},
  booktitle={International Conference on Learning Representations (ICLR)},
  year={2015}
}

@article{huang2006extreme,
  title={Extreme learning machine: Theory and applications},
  author={Huang, Guang-Bin and Zhu, Qin-Yu and Siew, Chee-Kheong},
  journal={Neurocomputing},
  volume={70},
  number={1--3},
  pages={489--501},
  year={2006},
  doi={10.1016/j.neucom.2005.12.126}
}

@article{huang2006universal,
  title={Universal approximation using incremental constructive feedforward networks with random hidden nodes},
  author={Huang, Guang-Bin and Chen, Lei and Siew, Chee-Kheong},
  journal={IEEE Transactions on Neural Networks},
  volume={17},
  number={4},
  pages={879--892},
  year={2006},
  doi={10.1109/TNN.2006.875977}
}

@inproceedings{rahimi2007random,
  title={Random features for large-scale kernel machines},
  author={Rahimi, Ali and Recht, Benjamin},
  booktitle={Advances in Neural Information Processing Systems},
  volume={20},
  pages={1177--1184},
  year={2007}
}

@inproceedings{le2013fastfood,
  title={{Fastfood}: Approximating kernel expansions in loglinear time},
  author={Le, Quoc V. and Sarl{\'o}s, Tam{\'a}s and Smola, Alex J.},
  booktitle={Proceedings of the 30th International Conference on Machine Learning (ICML)},
  volume={28},
  pages={244--252},
  year={2013}
}

@inproceedings{sutherland2015error,
  title={On the error of random {Fourier} features},
  author={Sutherland, Danica J. and Schneider, Jeff},
  booktitle={Proceedings of the 31st Conference on Uncertainty in Artificial Intelligence (UAI)},
  pages={862--871},
  year={2015}
}

@article{stanley2009hyperneat,
  title={A hypercube-based encoding for evolving large-scale neural networks},
  author={Stanley, Kenneth O. and D'Ambrosio, David B. and Gauci, Jason},
  journal={Artificial Life},
  volume={15},
  number={2},
  pages={185--212},
  year={2009},
  doi={10.1162/artl.2009.15.2.15202}
}

@inproceedings{chen2015hashednets,
  title={Compressing neural networks with the hashing trick},
  author={Chen, Wenlin and Wilson, James T. and Tyree, Stephen and Weinberger, Kilian Q. and Chen, Yixin},
  booktitle={Proceedings of the 32nd International Conference on Machine Learning (ICML)},
  volume={37},
  pages={2285--2294},
  year={2015}
}

@inproceedings{larsson2017fractalnet,
  title={{FractalNet}: Ultra-deep neural networks without residuals},
  author={Larsson, Gustav and Maire, Michael and Shakhnarovich, Gregory},
  booktitle={International Conference on Learning Representations (ICLR)},
  year={2017}
}

@inproceedings{ha2017hypernetworks,
  title={{HyperNetworks}},
  author={Ha, David and Dai, Andrew and Le, Quoc V.},
  booktitle={International Conference on Learning Representations (ICLR)},
  year={2017}
}

@inproceedings{vaswani2017attention,
  title={Attention is all you need},
  author={Vaswani, Ashish and Shazeer, Noam and Parmar, Niki and Uszkoreit, Jakob and Jones, Llion and Gomez, Aidan N. and Kaiser, {\L}ukasz and Polosukhin, Illia},
  booktitle={Advances in Neural Information Processing Systems},
  volume={30},
  pages={5998--6008},
  year={2017}
}

@inproceedings{li2018intrinsic,
  title={Measuring the intrinsic dimension of objective landscapes},
  author={Li, Chunyuan and Farkhoor, Heerad and Liu, Rosanne and Yosinski, Jason},
  booktitle={International Conference on Learning Representations (ICLR)},
  year={2018}
}

@inproceedings{frankle2019lottery,
  title={The lottery ticket hypothesis: Finding sparse, trainable neural networks},
  author={Frankle, Jonathan and Carbin, Michael},
  booktitle={International Conference on Learning Representations (ICLR)},
  year={2019}
}

@inproceedings{sitzmann2020siren,
  title={Implicit neural representations with periodic activation functions},
  author={Sitzmann, Vincent and Martel, Julien N. P. and Bergman, Alexander W. and Lindell, David B. and Wetzstein, Gordon},
  booktitle={Advances in Neural Information Processing Systems},
  volume={33},
  pages={7462--7473},
  year={2020}
}

@inproceedings{lin2020mcunet,
  title={{MCUNet}: Tiny deep learning on {IoT} devices},
  author={Lin, Ji and Chen, Wei-Ming and Lin, Yujun and Cohn, John and Gan, Chuang and Han, Song},
  booktitle={Advances in Neural Information Processing Systems},
  volume={33},
  pages={11711--11722},
  year={2020}
}

@misc{ma2024bitnet,
  title={The era of 1-bit {LLMs}: All large language models are in 1.58 bits},
  author={Ma, Shuming and Wang, Hongyu and Ma, Lingxiao and Wang, Lei and Wang, Wenhui and Huang, Shaohan and Dong, Li and Wang, Ruiping and Xue, Jilong and Wei, Furu},
  year={2024},
  howpublished={arXiv:2402.17764}
}

@misc{salimans2017evolution,
  title={Evolution strategies as a scalable alternative to reinforcement learning},
  author={Salimans, Tim and Ho, Jonathan and Chen, Xi and Sidor, Szymon and Sutskever, Ilya},
  year={2017},
  howpublished={arXiv:1703.03864}
}

@misc{sarkar2025eggroll,
  title={Evolution strategies at the hyperscale},
  author={Sarkar, Bidipta and Fellows, Mattie and Duque, Juan Agustin and Letcher, Alistair and Villares, Antonio Le{\'o}n and Sims, Anya and Wibault, Clarisse and Samsonov, Dmitry and Cope, Dylan and Liesen, Jarek and Li, Kang and Seier, Lukas and Wolf, Theo and Berdica, Uljad and Mohl, Valentin and Goldie, Alexander David and Courville, Aaron and Sevegnani, Karin and Whiteson, Shimon and Foerster, Jakob Nicolaus},
  year={2025},
  howpublished={arXiv:2511.16652}
}

@manual{stmicro2021stm32l,
  title={{STM32L476xx} Ultra-Low-Power {ARM Cortex-M4} Microcontroller Datasheet},
  organization={STMicroelectronics},
  year={2021},
  note={{DocID025976} Rev 8}
}

@misc{dagli2026fns_zenodo,
  title={Fractal Neural Synthesis: Source Code, Empirical Benchmarks, and Verification Suite},
  author={Da{\u{g}}l{\i}, Volkan and Da{\u{g}}l{\i}, Zerrin and Da{\u{g}}l{\i}, Da{\u{g}}han},
  year={2026},
  howpublished={Zenodo},
  doi={10.5281/zenodo.22774934},
  url={https://doi.org/10.5281/zenodo.22774934},
  note={Also archived at SSRN Abstract 7659198}
}

@misc{dagli2026werr_map,
  title={Universal Fractal Natural Language Decision Map: Real-Time Edge Triage Across Heterogeneous Domains},
  author={Da{\u{g}}l{\i}, Volkan and Da{\u{g}}l{\i}, Zerrin and Da{\u{g}}l{\i}, Da{\u{g}}han},
  year={2026},
  howpublished={arXiv:2609.25498}
}

@misc{dagli2026werracle,
  title={Werracle: Sub-Cent Intra-Block {AI} Reflex Oracles and Flash-Loan Circuit Breakers for {EVM} Smart Contracts},
  author={Da{\u{g}}l{\i}, Volkan and Da{\u{g}}l{\i}, Zerrin and Da{\u{g}}l{\i}, Da{\u{g}}han},
  year={2026},
  howpublished={arXiv:2609.30719}
}

@misc{dagli2026lean4_gauntlet,
  title={Zero-Storage Procedural Neural Synthesis via Boundary Dynamics: Formal Verification in {Lean 4} and Bare-Metal Gauntlet Validation},
  author={Da{\u{g}}l{\i}, Volkan and Da{\u{g}}l{\i}, Zerrin and Da{\u{g}}l{\i}, Da{\u{g}}han},
  year={2026},
  howpublished={arXiv:2609.33066}
}

\end{document}